\documentclass[11pt]{article}

\usepackage[margin=1in]{geometry}
\usepackage{amsmath,amssymb}
\usepackage{booktabs}
\usepackage{longtable}
\usepackage{array}
\usepackage{xcolor}
\usepackage{graphicx}
\usepackage{microtype}
\usepackage{tikz}
\usetikzlibrary{arrows.meta,calc}
\definecolor{caseL1}{HTML}{1F6FB2}
\definecolor{caseL2}{HTML}{2E8B57}
\definecolor{caseL3}{HTML}{C8102E}
\definecolor{caseL4}{HTML}{6A4C93}
\usepackage[numbers,sort&compress]{natbib}
\usepackage{authblk}
\usepackage[colorlinks=true,linkcolor=blue!50!black,citecolor=blue!50!black,urlcolor=blue!50!black]{hyperref}

\definecolor{phbg}{RGB}{251,234,234}
\definecolor{phfg}{RGB}{156,43,43}

\newcommand{\govprinciple}[1]{\par\medskip\noindent\textbf{Governance principle.} #1\par\medskip}

\title{\textbf{The CASE Framework: A Multi-Disciplinary Control Architecture for Governing Enterprise Agentic AI}}

\author[1]{Srinivas Telukunta}
\affil[1]{Cornell University, Johns Hopkins University}
\author[2]{Georgios Nektarios Lilis}
\affil[2]{Cornell University}
\author[3]{Lucio Baron}
\affil[3]{AI71}

\affil[ ]{\texttt{corresponding author email: st245@cornell.edu}}

\begin{document}
\maketitle

\begin{abstract}
Enterprises are deploying autonomous AI agents faster than they can govern them, and prevailing approaches stretch a single discipline, typically DevSecOps built for deterministic automation, across every scale of agency. We argue that agentic AI governance is four problems, not one, each with a mature governing science. The CASE framework assigns Control theory to the individual agent (intent as setpoint, guardrails as feedback, evaluation as observation), complex Adaptive systems theory to agent collectives (where emergence makes single-agent assurance non-compositional), Supervisory cybernetics to human-agent teams (where the Law of Requisite Variety shows unaided human oversight fails structurally), and Engineering operations to fleets (extending error budgets to decision quality so autonomy becomes a controlled variable). We formalize each layer, derive cross-layer coupling conditions, including a zero-touch deployment paradox where excellence at one layer strains the others, and trace twenty-plus enterprise controls to their classical constructs. Three empirical studies validate the thesis: 82 percent of documented production agent failures are multi-layer trajectories; none of 22 ecosystem tools offers full Layer 2 (emergence) coverage; and all 35 scored public deployments fall in the lowest maturity band. We name this mismatch, risk realized at the emergence layer against capability barely offered and practice absent, the Emergence Gap. A five-level maturity model with a non-compensatory bottleneck-weighted index and assessment instrument operationalizes CASE as a scientific rather than process maturity model, grounded in production enterprise agentic platforms. As EU AI Act Article 14 makes effective human oversight a legal requirement, only architectures satisfying requisite variety can make oversight real rather than ceremonial.
\end{abstract}

\noindent\textbf{Keywords:} AI governance, Agentic AI, Autonomous AI systems, AI agents, Multi-agent AI, LLM agents, AgentOps, Control theory, Complex adaptive systems, Supervisory cybernetics, Human-AI teaming, AI safety, Runtime governance, AI maturity model, multi-disciplinary governance

\section{Introduction}
Autonomous AI agents, systems that plan, reason, invoke tools, and execute multi-step workflows with limited or no human intervention, have moved from research demonstrations to production business operations within roughly three years. The defining property is autonomy itself: these systems decide for themselves how to pursue a goal, and can perform actions their designers never explicitly specified or intended. This capacity for self-directed decisioning beyond designed intent is what separates an agent from automation, and it is the root of the governance challenge this paper addresses. The governance apparatus has not kept pace. Industry surveys report that only about one in five enterprises has a mature governance model for autonomous agents, while a substantial share of agentic AI programs are projected to fail by 2027 due to inadequate governance and risk controls \citep{acharya2026}. The dominant response has been to extend familiar instruments: content guardrails, policy documents, review boards, and monitoring dashboards inherited from the machine learning operations era.

The root of the failure is a category confusion between automation and agency. Classical automation executes a fixed control flow over bounded outputs: its behavior is enumerable at design time, so the DevSecOps toolkit of release gates, static policy, and pipeline security is sufficient governance, because certifying the artifact certifies the behavior. Agents break that equivalence. An agent's behavior is a stochastic policy over an open action space, conditioned on inputs no test suite enumerates, adapted through context and memory, and coupled to other agents through shared state. Certifying the artifact no longer certifies the behavior, and the behavior now exists at four distinct scales: the individual agent pursuing a bounded task; the collective of interacting agents sharing memory, tools, and context; the human-agent team in which people supervise machine actors; and the automated fleet operating at a volume no human team can inspect. Each scale exhibits qualitatively different failure physics. A guardrail that stabilizes one agent says nothing about the cascade dynamics of two hundred interacting agents. A review board that approves a deployment says nothing about whether its members can absorb the behavioral variety the deployed system will generate. Governance frameworks that apply the automation toolkit across all four scales are, in effect, prescribing one medicine for four diseases.

Our central claim is that each scale of agency already has a governing science, developed and hardened over decades in other domains. Control theory has governed autonomous single systems since the servomechanism era \citep{wiener1948, astrom2008}. Complex adaptive systems theory has explained emergent collective behavior in markets, ecologies, and colonies since the 1990s \citep{holland1995, kauffman1993}. Supervisory cybernetics, anchored in the Law of Requisite Variety and the Viable System Model, has formalized the limits and architecture of human oversight since the 1950s \citep{ashby1956, beer1972}. Site reliability engineering has industrialized the operation of large stochastic systems since the 2000s \citep{beyer2016}. The CASE framework (\textbf{C}ontrol theory, complex \textbf{A}daptive Systems, \textbf{S}upervisory cybernetics, \textbf{E}ngineering operations) assigns each discipline to the scale it was built for.

The contributions of this paper are sixfold:
\begin{itemize}
  \item A four-layer governance framework, CASE, with an explicit mapping from scale of agency to governing discipline, and a formal apparatus for each layer.
  \item Cross-layer coupling conditions describing how failures propagate between scales, including a zero-touch deployment paradox in which Layer 4 excellence mechanically strains \mbox{Layers 2 and 3}.
  \item A mechanism inventory tracing more than twenty deployable enterprise controls to the specific classical constructs they implement, giving practitioners implementation guidance and auditors named artifacts.
  \item Empirical validation through three web-scale evidence studies: a failure taxonomy mapping, a tooling capability analysis, and an enterprise maturity distribution, whose headline findings are pervasive cross-layer coupling in the failure record and a universal Layer 2 governance void in both tooling and deployments.
  \item A five-level CASE maturity model with a non-compensatory composite index, delivered with a complete assessment instrument, anchored scoring bands, and a worked example (Appendix~\ref{app:instrument}).
  \item An enterprise illustration sample reference for Zero Touch Agent Deployment architecture, distilled from the operation of production agentic platforms across multiple global business functions at Fortune 50 scale.
\end{itemize}

The remainder of the paper proceeds as follows. Section~\ref{sec:lit} surveys the four literature streams and the emerging maturity-model literature, closing with the integration gap. Section~\ref{sec:novel} states the novel aspects of this work. Section~\ref{sec:framework} develops the framework, its formalism, its coupling conditions, and the mechanism inventory. Section~\ref{sec:validation} presents the empirical validation. Section~\ref{sec:maturity} introduces the maturity model, whose full assessment instrument appears in Appendix~\ref{app:instrument}. Section~\ref{sec:illustration} illustrates the framework through a reference Zero Touch Agent Deployment architecture. Section~\ref{sec:implications} draws strategic and regulatory implications, and Section~\ref{sec:limitations} discusses limitations before Section~\ref{sec:conclusion} concludes. Appendix~\ref{app:protocols} documents the coding protocols for the empirical studies.

\section{Literature Survey}\label{sec:lit}
Five bodies of literature bear on the governance of agentic AI. Four correspond to the CASE layers; the fifth, maturity modeling, addresses how organizations stage capability. We survey each stream from its classical foundations to its most recent application to LLM-based agents, and close with the synthesis that motivates this paper: each stream governs one scale of agency and is largely silent on the others.

\subsection{Control-theoretic approaches to individual agents}
The classical foundations are settled science. Feedback control of autonomous systems dates to Wiener-era servomechanisms \citep{wiener1948} and matured through state-space methods, with observability and controllability formalized by Kalman, stability by Lyapunov methods, receding-horizon constraint handling by model predictive control, and practical robustness devices such as gain scheduling and anti-windup compensation; standard treatments include \citet{astrom2008}. The transfer to LLM agents is recent and accelerating. One line of work reformulates LLM operator agents under advanced regulatory control theory, mapping each feedback loop to a specialized agent carrying explicit control-theoretic context: controlled variable, setpoint, chain priority, and selector logic, with conflicts resolved through structural priority rather than negotiation \citep{arc2026}. A second line arrives at control from the safety direction: guardrail research has converged on closed-loop designs in which guardrail feedback re-enters the planning context, outperforming binary allow-or-block filters and explicitly forming a feedback path between safety evaluation and action selection \citep{triad2026}. A third line embeds LLM reasoning inside classical control stacks for industrial plants, with action-simulation-validation cycles and safety overrides that transfer control to conservative fallback policies or human operators when iteration budgets are exhausted \citep{r2r2025, industrial2025}. Across all three lines the single-agent loop is the unit of analysis; none offers an account of what happens when controlled agents interact.

\subsection{Complex adaptive systems and multi-agent emergence}
Complex adaptive systems theory, developed by Holland, Kauffman, and the Santa Fe school \citep{holland1995, kauffman1993}, explains how local interaction rules among adaptive agents produce global order, phase transitions, and cascades that are irreducible to component behavior. Its adjacent apparatus is directly relevant to agent fleets: stigmergy explains coordination through a shared environment rather than direct messaging \citep{bonabeau1999}, self-organized criticality explains why coupled systems drift toward states where small perturbations trigger avalanches of all sizes \citep{bak1996}, and percolation theory relates graph connectivity to the reach of cascades. The framing has now been imported explicitly into the multi-agent LLM literature. Architecture taxonomies place complex adaptive systems at the highest level of multi-agent design, characterized by learning mechanisms, feedback loops, emergent behavior, and norm formation through decentralized interaction \citep{beyondstatic2026}. The empirical results are striking. Populations of 24 to 200 LLM agents in repeated coordination games spontaneously develop shared social conventions and collective biases present in no individual agent \citep{ashery2025}. Analyses of production frameworks such as AutoGen and ChatDev document bottom-up coordination under minimal supervision in which unspecified collective behaviors develop \citep{beyondstatic2026}. Reviews of large-scale deployments identify three structural properties that make such emergence hard to anticipate: irreducibility (isolated evaluation predicts little about population behavior), monoculture risk (shared base models correlate failures across nominally independent agents), and shared-state feedback through common memory and tools \citep{emergence2026}. The stream is descriptively rich but prescriptively thin: it explains why swarms surprise their designers without supplying the enterprise control machinery to bound the surprise.

\subsection{Cybernetics and the limits of human oversight}
The oldest and, we argue, most under-exploited stream. Ashby established that a regulator can hold an essential variable within bounds only if the variety of its responses matches the variety of the disturbances it must absorb, the Law of Requisite Variety \citep{ashby1956, ashby1958}. Conant and Ashby sharpened the requirement: every good regulator of a system must contain a model of that system \citep{conant1970}. Beer operationalized both results for organizations through the Viable System Model, allocating management capacity through recursive layers of variety attenuation and amplification, and adding two devices of direct relevance to agent oversight: the algedonic channel, a severity signal that bypasses the management hierarchy to reach accountable authority without filtering, and the recursion principle, under which every operational unit contains the full regulatory structure in miniature \citep{beer1972}. Sheridan formalized supervisory control for human operators of automation \citep{sheridan1992}, and second-order cybernetics extended the discipline to the observation of the observer, the question of who governs the governor. Application to agent oversight is now direct. Recent work argues from requisite variety that a point-in-time audit as the sole oversight mechanism is in the wrong variety class for a path-selecting agent fleet \citep{evidentiary2026}. Practitioner literature concedes the same point in plainer language: human review models collapse at agent speed and scale, forcing a redesign in which machines absorb variety while humans set boundaries and own consequences \citep{practitioner2026}. Regulation is forcing the issue from the other side. EU AI Act Article~14, enforceable for high-risk systems from August 2026, requires oversight by persons who can genuinely interpret, intervene, and override \citep{euaiact}, while a NIST-led agent standards initiative identifies opaque decision chains, emergent multi-agent behavior, and the practical impossibility of unaided real-time human oversight as the defining difficulties of the class \citep{nist2026}. What the stream lacks is integration with operational tooling: cybernetic requirements are stated, but the amplification machinery that could satisfy them is left unspecified.

\subsection{AgentOps and operational engineering}
Site reliability engineering codified the operation of large stochastic systems through service-level objectives, error budgets, golden signals, toil accounting, blameless postmortems, and, in its mature form, chaos engineering: deliberate fault injection to verify that resilience mechanisms fire in practice rather than on paper \citep{beyer2016}. AgentOps has emerged as its agent-native descendant because LLM-driven agents differ from traditional software in kind, not degree: services are provided by stochastic components, anomaly classes are broader, and root-cause analysis must reconstruct cognitive context rather than only system state \citep{agentops2025, agentops2024}. Systematic mappings catalogue a fast-growing tool ecosystem for tracing, evaluation, and cost control, organized around traces and spans that capture reasoning, plans, tool invocations, and guardrail status \citep{agentops2024}. SRE concepts transfer with meaningful adaptation: reliability for an agent includes decision quality, task completion fidelity, and cost-per-operation bounds, not merely availability, and tiered escalation architectures route incidents from autonomous recovery through AI-based investigation to human paging \citep{sreagents2026}. The stream is operationally rich but theoretically ungrounded: it accumulates practices without a principled account of why they work, where they will fail, or how operational telemetry relates to the control, emergence, and oversight problems of the other layers.

\subsection{Maturity models for agentic AI}
Several maturity models have appeared as enterprises seek staging discipline. The Agentic AI Governance Maturity Model defines five levels across twelve governance domains grounded in NIST AI RMF and ISO/IEC 42001, with simulation evidence linking higher maturity to lower agent sprawl and fewer risk incidents, and contributes a useful taxonomy of sprawl patterns including shadow agents, permission creep, and unmonitored delegation chains \citep{acharya2026}. Vendor models such as the Microsoft agentic AI adoption maturity model define progressive levels from experimentation to agent-first operation across strategy, governance, technology, and culture pillars \citep{msmaturity2026}, and consulting frameworks define autonomy ladders from copilots to orchestrated multi-agent ecosystems. These are \emph{process} maturity models: they measure whether governance activities exist and how broadly they are adopted. None asks the prior question of whether the discipline applied at each scale of agency is the correct one for the failure physics of that scale.

\subsection{Synthesis: what each stream governs and what it misses}
Table~\ref{tab:synthesis} summarizes the survey. Read column-wise, the literature is strong; read row-wise, it is fragmented. Control papers ignore emergence. Complex systems papers ignore the human variety budget. Oversight papers ignore operational scale. AgentOps lacks theory. Maturity models measure process rather than science. To our knowledge, no existing framework unifies the four disciplines with an explicit scale-to-discipline mapping, formal cross-layer coupling conditions, and a maturity pathway. That integration is the work of this paper.

\begin{table}[htbp]
\centering
\small
\caption{Literature synthesis: coverage and blind spots by stream.}
\label{tab:synthesis}
\begin{tabular}{p{3.4cm}p{3.0cm}p{3.6cm}p{3.6cm}}
\toprule
\textbf{Stream (key refs)} & \textbf{Scale it governs} & \textbf{Core instrument} & \textbf{Blind spot} \\
\midrule
Control theory \citep{wiener1948, astrom2008, arc2026, triad2026} & Individual agent loop & Feedback, observers, stability conditions & Non-compositional: silent on interaction effects \\
\addlinespace
Complex adaptive systems \citep{holland1995, kauffman1993, bak1996, ashery2025} & Agent collectives & Emergence, criticality, cascade analysis & Descriptive; little prescriptive control machinery \\
\addlinespace
Supervisory cybernetics \citep{ashby1956, beer1972, evidentiary2026} & Human-agent teams & Requisite variety, viable system architecture & Unintegrated with operational tooling \\
\addlinespace
AgentOps / SRE \citep{agentops2025, beyer2016, sreagents2026} & Automated fleet & Telemetry, SLOs, error budgets, escalation & Atheoretical; practices without failure physics \\
\addlinespace
Maturity models \citep{acharya2026, msmaturity2026} & Organizational adoption & Staged process capability & Measures process existence, not discipline fit \\
\bottomrule
\end{tabular}
\end{table}

\section{Novel Aspects of This Work}\label{sec:novel}
Against the surveyed literature, this paper makes seven claims to novelty. We state each with its nearest prior work so that the contribution boundary is explicit and auditable.

\paragraph{N1. The scale-to-discipline mapping itself.} Prior work applies one discipline to agentic AI at a time: control theory to single loops \citep{arc2026, triad2026}, complex systems language to swarms \citep{beyondstatic2026, emergence2026}, requisite variety to oversight \citep{evidentiary2026}, SRE to operations \citep{sreagents2026}. CASE is, to our knowledge, the first framework to assert and formalize that these four are jointly necessary and individually insufficient, with an explicit assignment of each discipline to the scale of agency whose failure physics it matches.

\paragraph{N2. Oversight as a testable engineering inequality.} The requisite variety condition of Equation~\eqref{eq:variety}, human variety times engineered amplification gain must meet or exceed agent behavioral variety at peak, converts Article 14-style oversight obligations from a governance narrative into a falsifiable design condition with named evidence artifacts (variety budgets, amplification chains, escalation SLAs). Prior applications of Ashby to AI governance are qualitative \citep{evidentiary2026, practitioner2026}; we make the inequality the certification object.

\paragraph{N3. Cross-layer coupling as the explanation for governance failure.} Section~\ref{sec:coupling} derives how perturbations at one layer shift the feasible region at adjacent layers, including the zero-touch deployment paradox: the better the fleet's deployment automation, the faster its interaction variety outruns its oversight capacity. This coupling argument explains a pattern the incident literature records but does not theorize: production failures are multi-layer trajectories, not single-layer events.

\paragraph{N4. A non-compensatory, scientific maturity index.} Existing maturity models average process scores \citep{acharya2026, msmaturity2026}. The CASE index is a deliberate consequence of N3: a bottleneck-weighted composite (Equation~\eqref{eq:index}) in which the weakest layer carries majority weight and the maturity level is read from the composite's band, with the geometric mean retained as the strict robustness limit. Coupled layers cannot compensate for one another, and the instrument yields a capital allocation rule (fund the binding layer first, because only that investment moves the minimum term) that arithmetic models cannot produce; Appendix~\ref{app:instrument} demonstrates this with a worked marginal analysis.

\paragraph{N5. Autonomy as a controlled variable.} We extend the SRE error budget from availability to decision quality and close the loop by modulating autonomy level against remaining budget (Equations~\eqref{eq:budget} and \eqref{eq:autonomy}). Prior AgentOps work measures decision quality \citep{agentops2025, sreagents2026}; making autonomy expansion and contraction a function of the measured budget is, to our knowledge, new as a stated governance mechanism.

\paragraph{N6. Validation methodology from public web evidence.} The three-study design of Section~\ref{sec:validation} (failure taxonomy mapping, tooling capability gap analysis, enterprise maturity distribution) validates a governance framework against the public record rather than against simulation alone \citep{acharya2026} or single-vendor telemetry. The method is replicable by any reader with access to the cited public sources, and the coding protocols are published in Appendix~\ref{app:protocols}.

\paragraph{N7. Theory-to-mechanism traceability.} The mechanism inventory of Section~\ref{sec:mechanisms} traces each deployable enterprise control to the specific classical construct it implements: receding-horizon planning to model predictive control, context trimming to anti-windup compensation, shared-memory governance to stigmergy, cascade monitoring to self-organized criticality, escalation bypass to Beer's algedonic channel, fault-injection drills to chaos engineering. This traceability is absent from both the practitioner literature, which names mechanisms without theory, and the theoretical literature, which names constructs without mechanisms. It gives implementers a principled reason for each control and gives auditors a named artifact to inspect for each theoretical requirement.

\medskip
An eighth element is positioning rather than novelty: the framework is grounded in the authors' sustained operation of production agentic systems at Fortune 50 scale (Section~\ref{sec:illustration}), so every governance principle in Section~\ref{sec:framework} corresponds to a mechanism the authors have run, not only proposed. Practitioners grounding of this kind remains rare in the agentic governance literature, which is dominated by simulation studies and vendor documentation.

\section{The CASE Framework}\label{sec:framework}
CASE rests on a single design rule: match the governing discipline to the scale of agency. Table~\ref{tab:mapping} summarizes the mapping; the subsections that follow formalize each layer, derive the coupling conditions, and then translate the formalism into a deployable mechanism inventory.

\begin{table}[htbp]
\centering
\small
\caption{The CASE scale-to-discipline mapping.}
\label{tab:mapping}
\begin{tabular}{lp{3.4cm}p{3.6cm}p{4.2cm}}
\toprule
\textbf{Layer} & \textbf{Scale of agency} & \textbf{Governing discipline} & \textbf{Characteristic failure} \\
\midrule
L1: C & Individual agent & Control theory & Drift, instability, unobservable state \\
L2: A & Agent collectives and swarms & Complex adaptive systems & Emergent cascades, contention, convention lock-in \\
L3: S & Human-agent teams & Supervisory cybernetics & Variety deficit, ceremonial oversight \\
L4: E & Automated fleet at scale & Engineering operations (SRE) & Silent degradation, unbounded cost, no rollback \\
\bottomrule
\end{tabular}
\end{table}

\subsection{Layer 1: Control theory for the individual agent}
Model the agent as a discrete-time controlled system, clocked not by wall time but by decision steps: unlike a physical plant sampled at fixed frequency, an agent loop is event-driven, and every construct that follows is native to discrete-time control, so nothing in the mapping requires a continuous, high-rate signal. Let $x_t$ denote the latent task state (plan, context, tool results), $u_t$ the action emitted by the agent policy, $w_t$ exogenous disturbance (adversarial input, tool failure, distribution shift), and $r_t$ the intent setpoint derived from the task specification:
\begin{equation}\label{eq:loop}
x_{t+1} = f(x_t, u_t, w_t), \qquad u_t = \pi(\hat{x}_t, r_t)
\end{equation}
where $\hat{x}_t$ is the state estimate produced by an observer implemented as tracing, evaluations, and telemetry. Three classical conditions become governance requirements. \emph{Observability}: if the observation function does not expose the components of state in which drift occurs, drift is undetectable by construction, so evaluation coverage is not a quality nicety but a structural precondition for control. \emph{Controllability}: the intervention surface (prompt revision, tool revocation, fallback policies, termination) must be able to drive the agent back into the safe operating region from any reachable state. \emph{Stability}: define a safety function $V(x)$ that is positive outside the certified operating envelope and require the closed loop to decrease it, the practical analogue of a Lyapunov condition, implemented as guardrail-enforced invariants with feedback into planning rather than binary blocking \citep{triad2026}.

Beyond the three conditions, four further constructs from the control canon transfer directly. Agent planning is naturally \emph{receding-horizon}: plan several steps, execute one, re-observe, re-plan under the constraint set, the structure of model predictive control, and governing the re-planning loop is what prevents open-loop plan execution from drifting. \emph{Gain scheduling} applies when a single guardrail strictness cannot fit all regimes: strictness should be scheduled by task risk class and input novelty, high-gain in unfamiliar regimes and relaxed in certified ones. \emph{Anti-windup} compensation applies when the action space saturates: accumulated refusals and failed attempts left in context integrate like windup in a saturated actuator, producing retry storms, so accumulated failure context must be trimmed or reset. And \emph{state estimation} in the Kalman tradition applies because no single evaluation signal is clean: multiple noisy eval signals should be fused into an estimate with explicit uncertainty, and the loop should respond to estimator variance, widening review when confidence is low.

\paragraph{Measurable instantiation.} The mapping is implementable, not only conceptual, because the error term is a vector of quantities production platforms already emit at each decision step. For a service-operations triage agent, a concrete error vector is: the guardrail verdict score for the proposed action (bounded, per step); the trajectory evaluation grade from a rubric-based judge against the declared intent (per step or per plan revision); semantic drift, the embedding distance between the current plan and the intent specification; the count of constraint violations attempted at the resource gateway; and budget burn rate against Equation~\eqref{eq:budget}. None of these is as clean as a thermocouple reading; each is noisy, which is precisely why Table~\ref{tab:mechanisms} pairs them with a Kalman-style observer that fuses them into a state estimate with explicit uncertainty and widens review when estimator variance is high. The controller acts on the fused estimate, exactly as classical loops act on filtered rather than raw signals.

\govprinciple{Every production agent runs inside a closed loop with an explicit setpoint, an observer whose coverage is audited against the state space, and an intervention surface proven able to reach a safe state. An agent without an observer is unmanaged by definition, whatever its guardrail count.}

\subsection{Layer 2: Complex adaptive systems for agent collectives}
Layer 1 assurance is non-compositional. For $n$ interacting agents sharing memory, tools, and message channels, collective risk decomposes as:
\begin{equation}\label{eq:swarm}
R_{\mathrm{swarm}} = \sum_{i=1}^{n} R_i + \sum_{i \neq j} \varphi(i, j) + \text{higher-order interaction terms}
\end{equation}
where $\varphi(i, j)$ captures pairwise interaction risk through shared state, tool contention, and communication. The interaction terms are precisely what single-agent evaluation cannot see, and empirical evidence shows they are not negligible: agent populations develop spontaneous conventions and collective biases absent from any member \citep{ashery2025}, and correlated base models create monoculture failure modes \citep{emergence2026}. Cascade dynamics supply the operative control variable. If an error in one agent triggers on average $k$ downstream errors, the collective is subcritical for $k < 1$ and supercritical above it:
\begin{equation}\label{eq:branching}
k = \mathbb{E}[\text{secondary failures per failure}]; \qquad \text{design target: } k < 1 \text{ under worst-case load}
\end{equation}

Three further constructs from the complex systems canon sharpen the layer. \emph{Stigmergy} \citep{bonabeau1999} identifies the shared environment, in agent fleets the common memory stores, vector indexes, and artifact repositories, as a coordination channel in its own right: agents coordinate, and contaminate, through what they write and read, so the medium itself must be governed with provenance tags, time-to-live, and write permissions. \emph{Self-organized criticality} \citep{bak1996} warns that coupled systems under continuous load drift toward critical states without any parameter being tuned there: the distribution of error-propagation avalanche sizes is the diagnostic, and a heavy tail is an early warning that the fleet is approaching criticality even while average behavior looks healthy. And \emph{percolation theory} relates the connectivity of the interaction graph to the reach of cascades: partitioning the graph into isolation cells keeps effective connectivity below the percolation threshold, bounding worst-case cascade size by cell size regardless of what emerges inside a cell. The ecological \emph{diversity-stability} principle adds the final instrument: heterogeneous base models on critical paths de-correlate the interaction terms that monocultures synchronize.

Enterprise instruments for holding $k$ below criticality therefore include interaction-level circuit breakers, rate and budget isolation between agent groups, diversity requirements on base models for critical paths, governed shared-memory media, avalanche-distribution monitoring, and registry-level dependency mapping so that the interaction graph is a managed artifact rather than an accident of integration.

\govprinciple{Collectives are certified at the interaction level, not the member level. The interaction graph, the branching factor under stress, and the shared-state channels are named, monitored, and bounded, and circuit breakers are placed where the graph analysis says cascades would propagate, not where org charts suggest.}

\subsection{Layer 3: Supervisory cybernetics for human-agent teams}
The Law of Requisite Variety states that only variety can absorb variety \citep{ashby1956, ashby1958}: a regulator holds an essential variable within bounds only if its response repertoire matches the disturbance repertoire it faces. In information terms, oversight capacity must satisfy:
\begin{equation}\label{eq:entropy}
H(O) \geq H(A) - H(A \mid O)
\end{equation}
where $H(A)$ is the entropy of agent behavior requiring regulation and $H(O)$ the variety of the oversight system. Since agent behavioral variety grows combinatorially with agent count, tool access, and autonomy level, while human cognitive variety is fixed, unaided human oversight fails mathematically, not managerially. Beer restated the law operationally as variety engineering \citep{beer1972}: oversight survives only through amplification of human variety and attenuation of agent variety. Writing $G$ for the engineered gain (automated triage, semantic summarization, tiered escalation, exception-only routing, AI-assisted supervision):
\begin{equation}\label{eq:variety}
V_{\mathrm{human}} \times G \geq V_{\mathrm{agents}}
\end{equation}
with the inequality required to hold at peak variety, not average load. This inequality reframes the regulatory debate. EU AI Act Article 14 requires oversight by persons able to interpret, intervene, and override \citep{euaiact}. Inequality~\eqref{eq:variety} states the engineering condition under which such oversight is genuine rather than ceremonial: without designed amplification, a human in the loop is a compliance artifact, and recent standards work acknowledges that real-time human oversight of long-running autonomous processes is otherwise practically impossible \citep{nist2026}. The Conant-Ashby theorem adds a second requirement: the oversight system must contain a model of the agents it regulates \citep{conant1970}, which in practice means supervisors operate on structured behavioral models (intents, envelopes, deviation scores), not raw traces.

Three further devices from the cybernetic canon complete the layer. Beer's \emph{algedonic channel} is a severity signal that bypasses the tier hierarchy entirely, reaching accountable authority without intermediate filtering; its agent-fleet analogue is a class of alerts that skip the escalation tiers and page the accountable executive directly, preventing the classic failure in which bad news attenuates as it climbs. Beer's \emph{recursion principle} holds that every viable operational unit contains the full regulatory structure in miniature; its analogue is federated supervision, in which each functional team runs the complete oversight loop for the agents it owns within enterprise-set envelopes, distributing oversight variety instead of concentrating it in a central bottleneck. And \emph{second-order cybernetics} demands that the oversight system observe itself: escalation SLA adherence, override drill outcomes, and approval-gate quality must themselves be measured, because an ungoverned governor fails silently.

\govprinciple{Oversight is an engineered channel with measured capacity. For every human-agent team, the organization computes peak agent variety, designs the amplification chain that closes inequality~\eqref{eq:variety}, and treats any gap as a hard deployment blocker, exactly as it would treat an unmet load requirement in physical infrastructure.}

\subsection{Layer 4: Engineering operations for automation at scale}
At fleet scale the governing science is site reliability engineering, extended from availability to decision quality \citep{beyer2016, sreagents2026}. Define an agent service-level objective over a decision population and derive an error budget:
\begin{equation}\label{eq:budget}
B = (1 - \mathrm{SLO}) \times N_{\mathrm{decisions}}
\end{equation}
where budget consumption counts hallucinated actions, scope violations, policy breaches, and cost overruns alongside conventional failures. The distinctive CASE move is to make autonomy itself a controlled variable modulated by remaining budget: expand autonomous scope while the budget is healthy, contract toward human review as it depletes, and halt expansion entirely on exhaustion:
\begin{equation}\label{eq:autonomy}
a(t) = g\big(B_{\mathrm{remaining}}(t)\big), \qquad g \text{ monotone non-decreasing}
\end{equation}

The mature SRE canon contributes four further practices, each adapted to agents. \emph{Chaos engineering} transfers as fault-injection drills against the fleet: staged agent terminations, poisoned tool responses, and simulated cascade triggers verify that Layer 1 through 3 mechanisms actually fire, converting resilience from documentation into demonstrated behavior. The \emph{golden signals} extend from latency, traffic, errors, and saturation to decision quality and cost per decision, so dashboards report what agents decided, not only whether they responded. \emph{Blameless postmortems} gain a CASE-specific step: every incident is coded to its primary and secondary layers, which feeds both organizational learning and the maturity assessment of Appendix~\ref{app:instrument}. And \emph{toil accounting} counts manual oversight actions as toil with explicit reduction targets, which converts Layer 3 amplification (raising the gain $G$) from an aspiration into a funded engineering objective.

Operationally this layer comprises the machinery that makes the other three layers cheap and repeatable: zero-touch deployment pipelines with policy gates, agent registries as systems of record, semantic observability on open telemetry standards, versioned rollback of prompts and policies as first-class release artifacts, and tiered incident response in which autonomous recovery, AI-based investigation, and human paging are explicit tiers with handoff contracts \citep{agentops2025, agentops2024, sreagents2026}.

\govprinciple{No agent reaches production except through the platform path, and autonomy is earned against a measured error budget rather than granted by approval memo. The platform, not the project team, owns the enforcement of Layers 1 through 3.}

\subsection{Cross-layer coupling and the zero-touch deployment paradox}\label{sec:coupling}
The layers are not independent, and their coupling explains why partial governance fails. A Layer 1 observability gap (unmeasured drift) raises the effective branching factor at Layer 2, because undetected errors propagate further before correction. A Layer 2 emergent behavior raises $H(A)$ at Layer 3, potentially breaking a variety inequality that was satisfied at design time. A Layer 3 variety deficit slows incident response, consuming Layer 4 error budget faster than modeled. Conversely, Layer 4 telemetry is the substrate on which Layer 1 observers, Layer 2 interaction maps, and Layer 3 amplification chains are all built. Formally, the safe operating condition is joint, not separable: perturbations $\delta$ at layer $l$ shift the feasible region at layers $l+1$ and $l-1$, so certification must be performed on the coupled system.

The coupling has a consequence sharp enough to deserve a name: the \emph{zero-touch deployment paradox}. Zero-touch deployment exists to remove the throughput bottleneck on agent creation, and it succeeds: the agent count $n(t)$ becomes a monotonically increasing function of time, limited by demand rather than by release engineering. But the interaction terms of Equation~\eqref{eq:swarm} grow as $O(n^2)$ in the worst case, and the behavioral variety $V_{\mathrm{agents}}$ of Inequality~\eqref{eq:variety} grows combinatorially with agent count, tool access, and shared channels. The oversight condition therefore becomes dynamic:
\begin{equation}\label{eq:ztad}
G(t) \geq \frac{V_{\mathrm{agents}}\big(n(t)\big)}{V_{\mathrm{human}}}, \qquad n(t) \text{ monotone increasing under zero-touch deployment}
\end{equation}
The better an organization's Layer 4, the faster it outruns a static Layer 3: deployment excellence mechanically manufactures the variety that oversight must absorb. This is not an argument against zero-touch deployment; it is the argument for why deployment automation and oversight amplification must be funded as a coupled pair, and it is the operational justification for the non-compensatory maturity index of Section~\ref{sec:maturity}. An organization that scores itself highly on deployment automation while its oversight gain $G$ is static is not maturing; it is accelerating toward the point where Inequality~\eqref{eq:ztad} fails.

The coupling is easiest to see in concrete, if composite, form. The following two scenarios are presented as possibilities assembled from failure modes the coded corpus of Section~\ref{sec:validation} documents individually, not as documented incidents. First, an IT-operations fleet runs thousands of agents for ticket triage, remediation, and identity workflows. One triage agent misclassifies a benign authentication anomaly as credential compromise, a Layer 1 estimation error. The classification is written to the shared context store that sibling agents consult, and stigmergic reuse turns one error into a fleet-level belief, Layer 2 contagion through the medium. Remediation agents act on the belief: account lockouts and device quarantines fan out, and the resulting alert storm arrives at the human tier as thousands of individually plausible tickets rather than one propagating misclassification, saturating oversight variety at Layer 3. Error budget burns at fleet scale while dashboards report every agent individually healthy at Layer 4. Every layer's mechanism inventory contains an interruption point: an uncertainty-aware observer widening review on a low-confidence classification, provenance tags on the shared store, an algedonic alert on the lockout-rate anomaly, a budget-triggered autonomy contraction. The trajectory is possible only where none of the four fires.

The second scenario concerns security, where two boundary statements sharpen the framework's scope. CASE presupposes conventional application security, authentication, hardening, penetration testing, and patching, as a precondition beneath Layer 1 rather than a governance layer within the framework; one of the two L3-primary incidents in Study 1, a classical remote-code-execution pivot, illustrates what happens when that floor is absent. The surface that does fall within scope, because it deliberately crosses layers, is untrusted input. A document ingested by a procurement agent carries an indirect prompt injection; the payload does not merely misdirect the ingesting agent (Layer 1) but instructs it to write poisoned entries into shared memory and to invoke a tool whose response is itself attacker-shaped (tool and MCP poisoning), recruiting sibling agents through the Layer 2 medium and generating escalation packets whose behavioral summaries launder the manipulation past Layer 3 review. Adversarial input is thus not a fifth layer but a stressor of all four at once, and it is why the population red-teaming and shared-memory governance mechanisms of Table~\ref{tab:mechanisms} treat the medium, and not only the agent, as the attack surface.

\subsection{Mechanism inventory: from constructs to controls}\label{sec:mechanisms}
The formalism above becomes actionable through named mechanisms. Table~\ref{tab:mechanisms} traces each deployable control to the classical construct it implements and the failure it prevents. Two entries, the model diversity policy at Layer 2 and behavioral-model escalation at Layer 3, extend the common practitioner inventory; both close gaps the literature explicitly documents (monoculture correlation \citep{emergence2026} and variety deficit disguised as information overload \citep{evidentiary2026}). The inventory also anchors the maturity scoring bands of Appendix~\ref{app:instrument}: layer maturity is assessed as evidence-backed coverage of these mechanism classes, which makes the rubric mechanically scoreable rather than judgment-only.

\begin{small}
\begin{longtable}{p{2.8cm}p{3.3cm}p{5.1cm}p{3.7cm}}
\caption{CASE mechanism inventory: each enterprise control traced to its classical construct.}\label{tab:mechanisms}\\
\toprule
\textbf{Mechanism} & \textbf{Classical construct} & \textbf{Implementation pattern} & \textbf{Failure prevented} \\
\midrule
\endfirsthead
\toprule
\textbf{Mechanism} & \textbf{Classical construct} & \textbf{Implementation pattern} & \textbf{Failure prevented} \\
\midrule
\endhead
\bottomrule
\endfoot
\multicolumn{4}{l}{\textbf{Layer 1: Control theory (stable, bounded agents)}} \\
\addlinespace
Closed-loop guardrails & Feedback control law (Eq.~\ref{eq:loop}) & Guardrail verdicts re-enter planning as proceed / refuse / revise signals & Silent goal drift across steps \\
\addlinespace
Receding-horizon planning & Model predictive control & Plan $N$ steps, execute one, re-observe, re-plan under the constraint set & Open-loop plan execution drift \\
\addlinespace
Budget caps & Reachable-set bounding & Token, cost, and wall-clock ceilings enforced at the gateway, not in the prompt & Runaway loops, unbounded cost \\
\addlinespace
Iteration limits with fallback & Finite-horizon Lyapunov condition & If $V(x)$ fails to decrease within $N$ iterations, control transfers to a fallback policy or human & Oscillation and thrash \\
\addlinespace
Anti-windup context management & Actuator saturation compensation & Trim accumulated failed-attempt context on repeated refusals & Retry storms, context poisoning \\
\addlinespace
Gain-scheduled guardrails & Gain scheduling & Strictness scheduled by task risk class and input novelty & Uniformly brittle or uniformly lax controls \\
\addlinespace
Uncertainty-aware observers & State estimation (Kalman tradition) & Fuse noisy eval signals into an estimate with confidence; widen review at high variance & Acting on a single noisy signal \\
\addlinespace
Least-privilege tool access & Action-space restriction (design-time variety attenuation) & Scoped per-agent credentials validated per call at the resource gateway & Injection-driven capability escalation \\
\addlinespace
Agent circuit breakers & Controllability, safe-state reachability & Kill-and-quarantine on invariant violation detected in telemetry & Compromised agent continues acting \\
\addlinespace
\midrule
\multicolumn{4}{l}{\textbf{Layer 2: Complex adaptive systems (ecosystem-level safety)}} \\
\addlinespace
Registry-backed interaction graph & Coupling structure of Eq.~\ref{eq:swarm} & Workflows declare shared-state channels at onboarding; undeclared edges block deployment & Invisible coupling through shared state \\
\addlinespace
Population simulation & Empirical estimation of branching factor $k$ & Staged replay of production traffic against the full declared graph & Member-certified, collective-untested \\
\addlinespace
Emergent behavior detection & Attractor and regime-shift analysis & Population metrics: output entropy, coordination indices, shared-vocabulary drift & Silent norm lock-in across the fleet \\
\addlinespace
Contagion controls & Percolation thresholds, graph partitioning & Rate and budget isolation cells; inter-cell breakers placed by graph analysis & Supercritical cascade ($k > 1$) \\
\addlinespace
Criticality monitoring & Self-organized criticality \citep{bak1996} & Avalanche-size distribution of error propagation; heavy-tail alerting & Drift toward critical state under healthy averages \\
\addlinespace
Shared-memory governance & Stigmergy \citep{bonabeau1999} & Provenance tags, time-to-live, and write permissions on shared context media & Artifact-mediated contamination \\
\addlinespace
Population red teaming & Worst-case perturbation analysis & Adversarial agents and poisoned shared state injected in simulation & Designing to average-case interaction risk \\
\addlinespace
Model diversity policy & Diversity-stability principle (ecology) & Heterogeneous base models or verified independent failure modes on critical paths & Monoculture cascade \\
\addlinespace
\midrule
\multicolumn{4}{l}{\textbf{Layer 3: Supervisory cybernetics (human-AI governance)}} \\
\addlinespace
Tiered autonomy & Variety attenuation \citep{ashby1956, beer1972} & Tier 0 autonomous absorption, tier 1 AI investigation, tier 2 human decision, with handoff SLAs & Oversight channel saturation \\
\addlinespace
Behavioral-model escalation & Good regulator theorem \citep{conant1970}; variety amplification & Escalation packets carry intent, deviation score, and proposed remediation, linked to full provenance & Variety deficit disguised as information overload \\
\addlinespace
Algedonic alerts & Beer's algedonic channel & Severity-triggered signals bypass tiers directly to accountable executives & Bad news filtered through hierarchy \\
\addlinespace
Approval gates & Consequence-weighted checkpoint placement & Risk-scored action classes; synchronous approval only for high-consequence classes & Rubber-stamping or ungated action \\
\addlinespace
Override authority and kill switches & Requisite intervention capacity (Article 14 operationalized) & Role-bound override rights; fleet and cell switches; quarterly drills with timing records & Ceremonial oversight \\
\addlinespace
Rewind, compensating rollback & Controllability extended to consequence space & Transactional and compensating-action patterns at the resource gateway & Irreversibility as the default \\
\addlinespace
Audit trails and provenance & Regulator's internal model requirement & Signed decision chains linking intent, context, tool calls, and outcomes & Unexplainable escalation \\
\addlinespace
Federated supervision & Beer's recursion principle (VSM) & Functional teams run the full oversight loop for owned agents within enterprise envelopes & Central oversight bottleneck \\
\addlinespace
Oversight meta-monitoring & Second-order cybernetics & Measure escalation SLA adherence, drill outcomes, and gate quality & Ungoverned governor \\
\addlinespace
\midrule
\multicolumn{4}{l}{\textbf{Layer 4: Engineering operations (zero-touch operationalization)}} \\
\addlinespace
Policy-as-code gates & Governance by construction & Pipeline verifies observer coverage, declared interaction edges, and tool scopes before admission & Shadow agents, permission creep \\
\addlinespace
Evals in CI/CD & Observability certified pre-release & Eval suites mapped to declared state space; coverage thresholds as gate criteria & Unobservable drift modes deployed \\
\addlinespace
Progressive rollout with atomic rollback & Error-budget-aware exposure scheduling & Canary cohorts; automatic reversion of code, prompts, and policies as one unit & Fleet-wide exposure to uncertified change \\
\addlinespace
Per-agent workload identity & Attribution precondition & Cryptographic identity at deployment; no shared credentials; identity-bound tool scopes & Unattributable action, credential sprawl \\
\addlinespace
Semantic observability & Shared telemetry substrate (Sec.~\ref{sec:coupling}) & Open-standard traces of reasoning, plans, tool calls, guardrail verdicts, and cost & Per-layer instrumentation silos \\
\addlinespace
Chaos and fault-injection drills & Chaos engineering \citep{beyer2016} & Staged agent kills, poisoned tool responses; verify Layer 1 to 3 mechanisms fire & Mechanisms that exist only on paper \\
\addlinespace
Extended golden signals & SRE golden signals & Latency, traffic, errors, saturation, plus decision quality and cost per decision & Availability-only dashboards \\
\addlinespace
Layer-coded postmortems & Blameless postmortem practice & Every incident coded to primary and secondary CASE layers; feeds maturity scoring & Unlearned incidents \\
\addlinespace
Error-budget-modulated autonomy & Eq.~\ref{eq:autonomy} closed loop & Budget dashboard by violation class; autonomy expansion approved against remaining budget & Autonomy granted by memo \\
\addlinespace
Toil accounting & SRE toil budgets & Manual oversight actions counted as toil; reduction targets fund amplification (raise $G$) & Permanent manual patchwork \\
\end{longtable}
\end{small}

\section{Empirical Validation}\label{sec:validation}
This framework extends beyond relabeling of established problems by offering a systematic basis for explaining, organizing, and predicting failures in autonomous AI systems. To assess its empirical validity, we evaluate three falsifiable claims using a broad corpus of publicly available evidence.

\textbf{Claim V1}: Documented production failures involving AI agents implicate all four CASE layers, either as primary failure mechanisms or as interacting secondary mechanisms. Moreover, a substantial proportion of these incidents involve mechanisms that cannot be adequately prevented or mitigated through Layer 1 controls alone.

\textbf{Claim V2}: The commercial and open-source tooling ecosystem is disproportionately concentrated on Layers 1 and 4, with comparatively limited capabilities addressing Layers 2 and 3. This structural imbalance indicates where enterprises are most likely to encounter governance, oversight, and operational blind spots.

\textbf{Claim V3}: Publicly documented enterprise deployments are predominantly concentrated at the lower levels of CASE maturity.

The evidence base was assembled exclusively from publicly available sources, including incident repositories, preprint archives, product documentation, engineering blogs, and published case studies. Sources were accessed in accordance with their applicable terms of use, and every coded item is referenced in the accompanying dataset. The coding methodology, inclusion criteria, and analytical protocols are detailed in Appendix~\ref{app:protocols}.

\subsection{Study 1: Failure taxonomy mapping}
\textbf{Method.} We assemble a corpus of documented agentic AI failures from public incident databases (the AI Incident Database and the MIT AI Risk Repository), vendor and platform postmortems, field studies published by observability providers, and peer-reviewed failure-mode analyses. Inclusion criteria: the system exhibits autonomous multi-step behavior (planning plus tool execution), the failure is described with enough mechanism detail to code, and the account is independent or self-reported with technical specifics. Each incident is coded to a primary CASE layer using the decision protocol of Appendix~\ref{app:protocols}, which also states the corpus scope rules (physical robotic autonomy and benchmark-only studies are excluded). Dual machine coding with first-author adjudication yields an inter-rater reliability statistic; Appendix~\ref{app:protocols} documents the exact coding configuration. Secondary layers are recorded to surface the coupling patterns of Section~\ref{sec:coupling}.

\begin{figure}[htbp]
\centering
\includegraphics[width=0.62\linewidth]{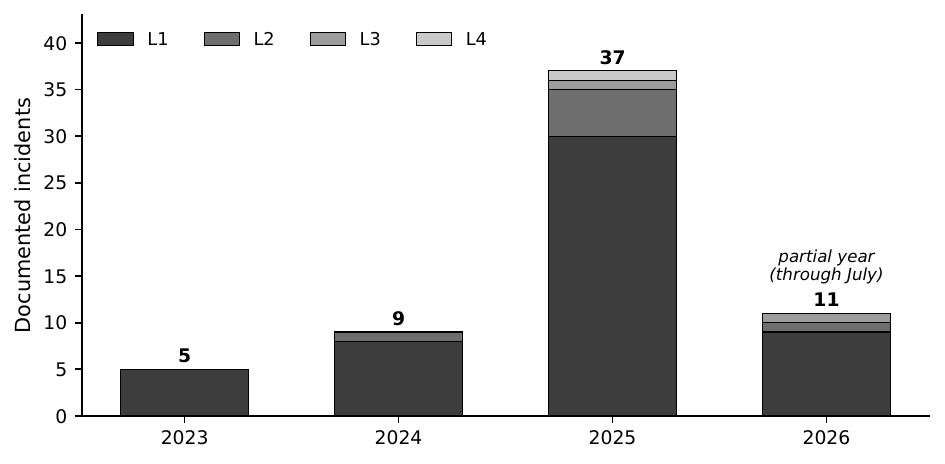}
\caption{The failure corpus over time: documented agentic AI incidents by year and primary CASE layer ($N=62$; 2026 covers January through July). The record is young and accelerating, and Layer 2 primaries appear only from 2024 onward, as multi-agent deployments begin to enter production.}
\label{fig:timeline}
\end{figure}

\textbf{Results.} The corpus comprises $N=62$ coded incidents (35 from the AI Incident Database, 26 from vendor postmortems and researcher disclosures, 1 from the MIT AI Risk Repository), and it is young: 5 incidents date from 2023, 9 from 2024, 37 from 2025, and 11 from the first seven months of 2026 (Figure~\ref{fig:timeline}), so more than three quarters of the documented record accumulated in the eighteen months preceding this study, consistent with the leading-indicator reading of Section~\ref{sec:outcomes}, coded twice under an intra-protocol consistency check, two differently worded prompts to the same model, with first-author adjudication; pre-adjudication agreement was Cohen's $\kappa=0.83$ (16 disagreements across 363 screening units), a measure of protocol robustness rather than coder independence. An independent cross-provider check, applying the identical, unreworded coder-A protocol prompt to a second model from a different provider (GPT-5.6-terra) across all 363 screened units, yields Cohen's $\kappa=0.643$ before adjudication (32 disagreements); adjudicating those disagreements added 13 incidents the intra-protocol process had excluded and reassigned one primary code; one admission, a physical autonomous-vehicle incident, was subsequently reversed on scope review for consistency with the corpus rules of Appendix~\ref{app:protocols}, yielding the corpus reported below (Appendix~\ref{app:protocols} reports the full cross-model agreement statistics). Table~\ref{tab:failures} reports the primary-layer distribution. The striking pattern is not that the L2 plus L3 primary share is large --- it is $15\%$ (9 of 62) --- but that a single primary code understates each incident: $51$ of $62$ ($82\%$) carry at least one secondary layer, and the coupling runs downward from the control layer. Of the $52$ incidents coded primarily to L1, $36$ ($69\%$) also implicate a supervisory (L3) mechanism and $43$ ($83\%$) an engineering-operations (L4) mechanism as a secondary code (Table~\ref{tab:cooccur}). The coupling is not confined to L1 primaries: $P(\mathrm{L3}\,\vert\,\mathrm{L2}) = 57\%$ (4 of 7) and $P(\mathrm{L4}\,\vert\,\mathrm{L2}) = 43\%$ (3 of 7), and both L3-primary incidents carry an L4 secondary. Averaged over the corpus, each incident implicates $2.61$ CASE layers, and the headline proportions carry 95\% Wilson intervals: $[73, 91]$ for the $84\%$ L1 primary share, $[71, 90]$ for the $82\%$ multi-layer share, and $[8, 25]$ for the $15\%$ L2-plus-L3 primary share.

\begin{table}[htbp]
\centering
\small
\caption{Distribution of $N=62$ documented agentic AI failures across CASE primary layers (reliability: intra-protocol $\kappa=0.83$, cross-model $\kappa=0.643$, all disagreements first-author adjudicated). Counts, shares, and representative failure modes (the most frequent coded mechanism phrases per layer) are emitted by the coding pipeline into \texttt{results/study1\_table.tex} and reproduced here.}
\label{tab:failures}
\begin{tabular}{p{2.6cm}p{7.2cm}cc}
\toprule
\textbf{Primary layer} & \textbf{Representative failure modes observed} & \textbf{Count} & \textbf{Share} \\
\midrule
L1 Control & indirect prompt injection exfiltration; injected-prompt fund transfer; jailbreak prompt abuse & 52 & 84\% \\
L2 Adaptive systems & orchestrator-driven multi-persona coordination; remote-confirmation request backlog; cross-agent prompt-injection relay & 7 & 11\% \\
L3 Supervisory & system-prompt injection contamination; unauthenticated rce exploitation pivot & 2 & 3\% \\
L4 Engineering ops & audit-log evasion via prompt & 1 & 2\% \\%
\bottomrule
\end{tabular}
\end{table}

\begin{table}[htbp]
\centering
\small
\caption{Secondary-code co-occurrence across the $N=62$ incidents: for each primary layer, the number of incidents also carrying each layer as a secondary code. Fifty-one incidents (82\%) carry at least one secondary code, the quantitative signature of the cross-layer coupling of Section~\ref{sec:coupling}.}
\label{tab:cooccur}
\begin{tabular}{lcccc}
\toprule
\textbf{Primary layer ($n$)} & \textbf{L1 secondary} & \textbf{L2 secondary} & \textbf{L3 secondary} & \textbf{L4 secondary} \\
\midrule
L1 Control (52) & --- & 10 & 36 & 43 \\
L2 Adaptive systems (7) & 0 & --- & 4 & 3 \\
L3 Supervisory (2) & 0 & 0 & --- & 2 \\
L4 Engineering ops (1) & 1 & 0 & 1 & --- \\
\bottomrule
\end{tabular}
\end{table}

\textbf{Interpretation.} We had anticipated that a material share of failures would carry a \emph{primary} L2 or L3 code; the data do not bear this out, and the reason is instructive. Under the coding protocol's earliest-layer rule (Appendix~\ref{app:protocols}), an incident is assigned to the first layer whose correct functioning would have interrupted the failure trajectory, so any failure a well-designed single-agent loop could also have caught is recorded as L1 --- including most oversight breakdowns, where a least-privilege boundary or circuit breaker would have contained the same event. The primary distribution therefore compresses toward L1 by construction. The evidence for Claim V1 lies instead in the co-occurrence structure: the failures that dominate the corpus are not \emph{single-layer} events. Governance that hardens only the control loop (L1) confronts a population in which $68\%$ of those very L1 incidents also implicate a supervisory (L3) mechanism and $83\%$ an engineering-operations (L4) mechanism --- exactly the layers a guardrail-centric stack does not instrument. Single-discipline governance thus leaves entire failure \emph{classes} unaddressed not because L2/L3 failures are common as primary codes, but because the L1 failures it does address are coupled to L3 and L4 mechanisms it does not. The full secondary-code co-occurrence matrix (Section~\ref{sec:coupling}) makes the coupling explicit. Claim V1 is therefore supported in its coupled form: $83\%$ of incidents carry failure mechanisms that Layer~1 instruments alone do not reach, even where a Layer~1 mechanism could have interrupted the trajectory. Counting involvement rather than primaries makes the point directly: a layer is implicated, as primary or secondary, in 85\% (L1), 27\% (L2), 69\% (L3), and 79\% (L4) of incidents, so at the mechanism level the public failure record is nearly as much a Layer 3 and Layer 4 record as a Layer 1 record (Figure~\ref{fig:synthesis}A).

\subsection{Study 2: Tooling capability gap analysis}
\textbf{Method.} We enumerate the agent-operations tool ecosystem from published systematic mappings \citep{agentops2024}, public repository metadata, and vendor documentation, then code each tool's documented capabilities against the four layers, using the mechanism classes of Table~\ref{tab:mechanisms} as the coding frame: does the tool implement any L1 mechanism class, any L2 class, any L3 class, any L4 class. Coding uses documentation evidence only, avoiding vendor claims without described mechanisms.

\textbf{Results.} We coded 22 agent-operations tools spanning three categories --- guardrails and gateways (6), observability and evaluation platforms (9), and orchestration and human-in-the-loop runtimes (7) --- against the mechanism classes of Table~\ref{tab:mechanisms}, Full / Partial / None per layer, with one line of documentation evidence recorded per non-None cell in the published dataset. Table~\ref{tab:tooling} gives the matrix; aggregate Full-coverage shares are L1 $50\%$, L2 $0\%$, L3 $36\%$, L4 $41\%$. Two findings are decisive. First, \emph{no tool in the sample provides Full L2 coverage} and only three provide even Partial: nothing on the market monitors the inter-agent interaction graph, cascade propagation, or shared-state contamination that Layer 2 names. Second, coverage is \emph{segregated by tool category}: guardrails and gateways cover L1 ($67\%$ Full) and essentially nothing else ($0\%$ on L2/L3/L4); observability platforms cover L4 ($89\%$ Full) and little else; and Full L3 oversight appears in every one of the orchestration/HITL runtimes yet is absent ($0$--$11\%$) from the guardrails and observability tooling that dominates enterprise adoption. No category covers all four layers; Table~\ref{tab:toolcat} quantifies the segregation. Counting Partial coverage as well does not close the Layer 2 gap: $86\%$ of tools show at least partial L1 capability, $82\%$ L3, and $91\%$ L4, against $14\%$ for L2. The absence is not sampling noise: comparing Full coverage at L2 (0 of 22) with L1 (11 of 22) gives a one-sided Fisher exact $p = 9.2 \times 10^{-5}$. The ecosystem is also fragmented by breadth: 2 of the 22 tools cover exactly one layer at Full-or-Partial, 5 cover two, 12 cover three, and only 3 cover all four, so 19 of 22 leave at least one layer wholly uncovered, and in all 19 the uncovered layer is, or includes, L2.

\begin{table}[htbp]
\centering
\small
\caption{Tool-by-layer capability matrix: 22 agent-operations tools coded Full / Partial / None on whether their public documentation at the time of this study evidences at least one CASE mechanism class per layer (config and per-cell evidence in the published dataset). Emitted by the documentation-coding pipeline into \texttt{results/study2\_table.tex} and reproduced here.}
\label{tab:tooling}
\begin{tabular}{p{4.2cm}cccc}
\toprule
\textbf{Tool / platform} & \textbf{L1 Control} & \textbf{L2 Emergence} & \textbf{L3 Oversight} & \textbf{L4 Fleet ops} \\
\midrule
LangSmith & Partial & None & Partial & Full \\
Langfuse & None & None & Partial & Full \\
Arize Phoenix & Partial & None & None & Full \\
Braintrust & None & None & None & Full \\
Datadog LLM Observability & Full & None & Partial & Full \\
HoneyHive & None & None & Partial & Full \\
AgentOps & Partial & None & None & Partial \\
Fiddler AI & Partial & Partial & Partial & Full \\
Orq.ai & Partial & Partial & Full & Full \\
Guardrails AI & Partial & None & None & None \\
NVIDIA NeMo Guardrails & Full & None & Partial & Partial \\
Lakera Guard & Full & None & Partial & Partial \\
Portkey & Full & None & Partial & Partial \\
LiteLLM & Full & None & Partial & Partial \\
Invariant Labs (Guardrails/Explorer) & Partial & Partial & Partial & Partial \\
LangGraph & Full & None & Full & None \\
CrewAI & Full & None & Full & Partial \\
Microsoft AutoGen & Full & None & Full & Partial \\
Temporal (durable execution for agents) & Full & None & Full & Partial \\
OpenAI Agents SDK & Full & None & Full & Partial \\
HumanLayer & Partial & None & Full & Partial \\
Azure AI Foundry Agent Service & Full & None & Full & Full \\
\midrule
\textbf{Column coverage (Full)} & 50\% & 0\% & 36\% & 41\% \\%
\bottomrule
\end{tabular}
\end{table}

\begin{table}[htbp]
\centering
\small
\caption{Full-coverage share by tool category and CASE layer. Each category is strong at the one or two layers its parent discipline owns and blind elsewhere; no category, and no tool, provides Full Layer 2 coverage.}
\label{tab:toolcat}
\begin{tabular}{lcccc}
\toprule
\textbf{Tool category ($n$)} & \textbf{L1} & \textbf{L2} & \textbf{L3} & \textbf{L4} \\
\midrule
Guardrails and gateways (6) & 67\% & 0\% & 0\% & 0\% \\
Observability and evaluation (9) & 11\% & 0\% & 11\% & 89\% \\
Orchestration and HITL runtimes (7) & 86\% & 0\% & 100\% & 14\% \\
\midrule
All tools (22) & 50\% & 0\% & 36\% & 41\% \\
\bottomrule
\end{tabular}
\end{table}

\textbf{Interpretation.} The market corroborates the framework through its own segmentation. An enterprise assembling the mainstream stack --- a guardrail layer in front of the model plus an observability and evaluation platform behind it --- acquires strong Layer 1 and Layer 4 instruments and almost no Layer 2 or Layer 3 capability. Interaction-graph and cascade monitoring (Layer 2) do not exist as products in the sample; oversight instrumentation (Layer 3) ships only inside dedicated orchestration runtimes, which a monitoring-first buyer may never adopt and which are not where enterprises watch their fleets. The single-discipline pattern this paper identifies in the failure record reappears at the tooling layer: each tool category is strong at the one or two layers its discipline owns and blind to the rest, and no product spans all four. The gap doubles as a roadmap --- Layer 2 monitoring is a green field, and Layer 3 oversight needs to migrate out of the orchestration frameworks and into the observability plane where agents are actually watched. Claim V2 is supported: the monitoring and guardrail tooling enterprises deploy is structurally overweight Layers 1 and 4, carries no Layer 2 capability at all, and carries Layer 3 capability only if a dedicated orchestration runtime is adopted alongside it.

\subsection{Study 3: Enterprise maturity distribution}
\textbf{Method.} We score publicly describable enterprise agent deployments against the CASE maturity instrument of Appendix~\ref{app:instrument} using published case studies, engineering blog disclosures, conference talks, and hiring signals (the oversight and platform roles an organization advertises reveal the layers it operates). Each deployment receives four layer scores from documented evidence, conservative-coded: absence of evidence scores as absence of capability, which biases the study against our hypothesis only if enterprises systematically under-disclose mature practice.

\begin{figure}[htbp]
\centering
\includegraphics[width=\linewidth]{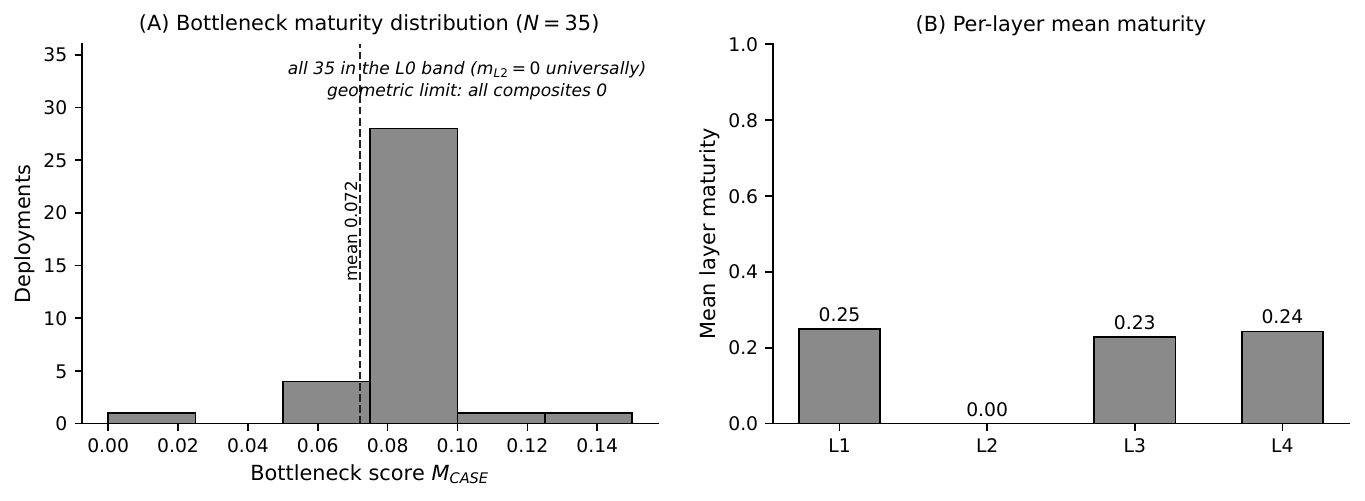}
\caption{CASE maturity across $N=35$ public enterprise agent deployments (conservative-coded on the Appendix~\ref{app:instrument} bands). (A)~Histogram of the bottleneck index $M_{\mathrm{CASE}}$ (Equation~\eqref{eq:index}, $\alpha=0.6$): scores spread from $0.000$ to $0.125$, yet every composite falls in the L0 band because the emergence layer is universally ungoverned; under the strict geometric limit all composites are exactly zero. (B)~Per-layer mean maturity: partial coverage at L1, L3, and L4, and a categorical zero at L2. Generated by the scoring pipeline into \texttt{results/figure1.pdf}.}
\label{fig:maturity}
\end{figure}

\textbf{Results.} We scored $N=35$ publicly describable enterprise agent deployments spanning three sectors (big-tech and SaaS platforms, financial and professional services, and customer-support and vertical applications) on the anchored bands of Appendix~\ref{app:instrument}, conservative-coded, with one line of documentary evidence recorded per non-zero layer in the published dataset (Figure~\ref{fig:maturity}). The distribution is even more concentrated than we hypothesized: \emph{every} deployment falls in the composite L0 band. The reason is structural. Per-layer mean maturities are L1 $0.25$, L2 $0.00$, L3 $0.23$, and L4 $0.24$: enterprises publicly evidence partial control-layer (L1) and operations-layer (L4) capability and some supervisory (L3) capability, but not one of the 35 evidences a single Layer~2 (emergence) mechanism class. Under the bottleneck composite of Section~\ref{sec:index}, that universal zero holds every deployment in the L0 band: all 35 composites fall below the first band threshold by Equation~\eqref{eq:index}, spreading from $0.000$ to $0.125$ (mean $0.072$, median $0.075$), separating deployments with partial L1, L3, and L4 capability from those evidencing none. The anchored bands quantize the score, and the quantization is itself a finding: 28 of the 35 deployments share one identical evidence signature, band one ($0.25$) at L1, L3, and L4 with zero at L2, so the modal enterprise agent deployment in the public record is not merely low-maturity but uniform. The signature is also sector-invariant: big-tech and SaaS platforms, financial and professional services, healthcare, retail, telecom, and public-sector deployments all exhibit it, including the universal Layer 2 zero, so the emergence gap is a property of current practice, not of any one industry. Conservative-coded absence across 35 independent deployments also bounds the claim statistically: by the rule of three, the true prevalence of any deployed L2 mechanism in the population this sample represents is below $8.6\%$ at 95\% confidence. The distance to the next level is equally concentrated: for 30 of the 35 deployments, Layer 2 is the only layer below the first band, so a single band of L2 evidence ($0.25$) would lift their composites exactly to the L1 threshold, making build-L2 not only the highest-leverage investment but the cheapest level move in the population. Under the strict geometric robustness limit, every composite is exactly zero. The modal weakest layer is accordingly L2, weakest or tied weakest in all 35 deployments; L3 ties as weakest in 4 deployments, L4 in 2, and L1 in 1.

\textbf{Interpretation.} We hypothesized a distribution concentrated at L0 to L1 with Layer 3 as the modal weakest layer, consistent with survey findings that mature agent governance remains rare \citep{acharya2026}. The hypothesis of low maturity concentrated at the bottom of the scale is supported, and then some: not merely L0 to L1 but a complete L0 pileup. The specific sub-prediction that Layer 3 would be the modal weakest layer is not borne out; Layer 2 is, because emergence-layer governance is absent from the entire public sample, a finding that echoes Study 2, where no tool implements Layer 2 monitoring either. The result is also the sharpest available demonstration of why the index is deliberately non-compensatory: an arithmetic average of the same scores reports a mean of $0.18$, an ``emerging'' L0-to-L1 population, papering over the fact that every one of these deployments runs multi-agent or tool-using systems with zero interaction-layer governance. Under the coupling conditions of Section~\ref{sec:coupling} that is not a rounding difference; it is the difference between a population that looks modestly immature and one that is uniformly fragile. The bottleneck score preserves discrimination inside that verdict: the population can be ordered and tracked from $0.000$ to $0.125$ even while every member sits in the same band. Claim V3 is supported. The universal-L0 result also needs interpretation rather than alarm: it does not mean the bar is set where nobody can reach it. The L1 band begins at $M_{\mathrm{CASE}} = 0.25$, a deliberately modest floor, and the modal deployment already holds band-one evidence at three of the four layers; the population fails on exactly one, and 30 of 35 deployments would reach the L1 threshold with a single band of Layer 2 evidence, concretely, a registry-backed interaction graph and one population-level metric in production. The finding is therefore not that the industry can do nothing, but that it is uniformly not doing one specific thing, and meaningful progress has a precise, modest definition. The conservative-coding caveat of Section~\ref{sec:threats} applies with force: public disclosure rarely substantiates the measured-coverage evidence the higher bands require, so these scores are lower bounds, and the per-layer profile identifies where the enterprise population is systematically exposed as regulators approach the Article 14 enforcement date. The Layer 2 gap, however, is a gap in kind rather than degree, and is unlikely to be a disclosure artifact given that, as Study 2 shows, the tooling to close it does not yet exist.

\subsection{Summary: hypotheses and outcomes}\label{sec:outcomes}
Table~\ref{tab:hypotheses} states each hypothesis as registered before coding, the observation, and the verdict, including the two sub-predictions the data rejected. Both misses point in the same direction: the emergence layer is emptier, in practice and in tooling, than even this framework anticipated, and primary-code accounting understates how strongly failures couple across layers. The framework's weakest-layer logic thus survives its own mispredictions; what the data corrected was our estimate of \emph{which} layer is weakest and \emph{where} the multi-layer signal shows up.

\begin{figure}[htbp]
\centering
\includegraphics[width=\linewidth]{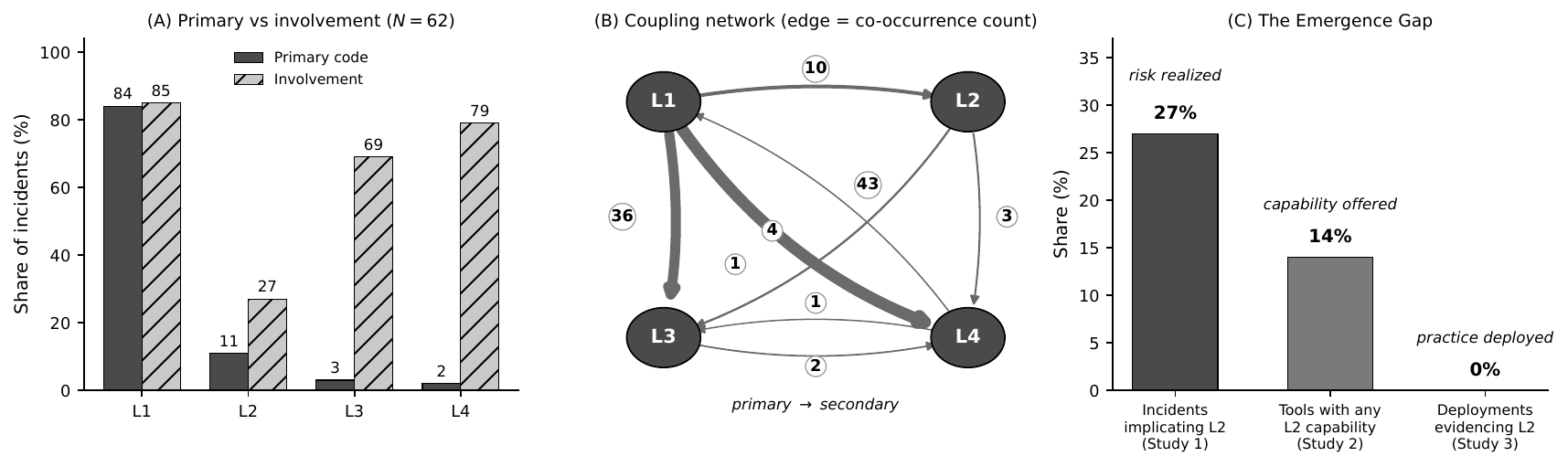}
\caption{Cross-study synthesis. (A)~Two accountings of the $N=62$ failure corpus: the share of incidents in which each layer is the primary code versus the share in which it is implicated at all (primary or secondary). Primary accounting says L1; involvement accounting says the record is multi-layer. (B)~The coupling network: nodes are CASE layers, directed edges weighted by primary-to-secondary co-occurrence counts. (C)~The Emergence Gap triangulated: share of incidents implicating L2 mechanisms (Study 1), share of tools offering Full or Partial L2 capability (Study 2), and share of deployments evidencing any L2 mechanism (Study 3).}
\label{fig:synthesis}
\end{figure}

Figure~\ref{fig:synthesis} draws the three studies together. Panel~A contrasts the two accountings of the failure record: primary codes concentrate at L1, but involvement (primary or secondary) puts L3 and L4 within striking distance of L1, which is the coupling thesis in a single chart. Panel~B renders the coupling structure as a network: the L1-to-L4 and L1-to-L3 edges dominate (44 and 36 incidents), confirming that when a single-agent loop fails, the operational and supervisory mechanisms around it were failing too. Panel~C triangulates what we name the \emph{Emergence Gap}: the systemic mismatch between the frequency with which collective interaction contributes to failures and the near absence of governance that manages emergence. Layer 2 mechanisms are implicated in 27\% of documented incidents, yet only 14\% of tools offer even partial Layer 2 capability, and not one of the 35 scored deployments evidences any; L2 is moreover the weakest or tied-weakest layer in every single deployment. Risk is being realized at a layer for which capability is barely offered and practice is entirely absent; that mismatch, more than any single statistic, is the central empirical result of this paper. One reading of the Gap deserves preemption: the tooling void (Study 2) and the practice void (Study 3) may be one fact observed twice, since few enterprises yet operate large multi-agent estates, so there are no buyers and hence no tools. We accept that common-cause structure between the two studies and note that the Gap's third leg is independent of it: Layer 2 mechanisms are already implicated in 27\% of documented incidents at today's adoption levels. The Emergence Gap is therefore both a present risk and a leading indicator, and it will bind hardest exactly when multi-agent adoption arrives at the tooling and practice it currently lacks.

Read jointly, the three studies do more than rank layers; they diagnose a \emph{different} gap at each (Table~\ref{tab:gaps}). At L1 the gap is coupling: controls exist, are widely sold, and are partially deployed, yet 68\% and 83\% of L1-primary failures implicate L3 and L4 mechanisms alongside them. At L2 the gap is capability: risk is realized in 27\% of incidents against 0\% Full tooling and 0\% deployed practice. At L3 the gap is depth: 82\% of tools claim some oversight capability, but nearly half of those claims are Partial (10 of 22 tools), and Full mechanisms ship only inside orchestration runtimes. At L4 the gap is adoption: the tooling is the most mature of the four layers (91\% Full or Partial), yet deployed practice averages $0.24$. The remedy therefore differs by layer, build (L2), deepen (L3), adopt (L4), and couple (L1), a more actionable prescription than any single composite could carry.

\begin{table}[htbp]
\centering
\small
\caption{Gap diagnosis by layer, synthesizing the three studies: share of incidents in which the layer is implicated (Study 1), tooling coverage (Study 2), mean deployed practice (Study 3), and the distinct gap each layer exhibits.}
\label{tab:gaps}
\begin{tabular}{lccccp{4.2cm}}
\toprule
\textbf{Layer} & \textbf{Involvement} & \textbf{Tooling Full} & \textbf{Tooling Partial} & \textbf{Practice $m_l$} & \textbf{Diagnosed gap} \\
\midrule
L1 Control & 85\% & 50\% & 36\% & 0.25 & Coupling: controls fail jointly with L3 and L4 mechanisms \\
\addlinespace
L2 Adaptive systems & 27\% & 0\% & 14\% & 0.00 & Capability: no tooling, no practice, risk already realized \\
\addlinespace
L3 Supervisory & 69\% & 36\% & 45\% & 0.23 & Depth: oversight claimed widely, mechanism detail thin \\
\addlinespace
L4 Engineering ops & 79\% & 41\% & 50\% & 0.24 & Adoption: mature tooling, lagging deployed practice \\
\bottomrule
\end{tabular}
\end{table}

\begin{table}[htbp]
\centering
\small
\caption{Hypotheses registered ex ante versus outcomes observed across the three studies.}
\label{tab:hypotheses}
\begin{tabular}{p{2.5cm}p{3.6cm}p{4.6cm}p{3.0cm}}
\toprule
\textbf{Hypothesis} & \textbf{As stated ex ante} & \textbf{Observed} & \textbf{Verdict} \\
\midrule
V1a: primary distribution & Material share of primary codes at L2 or L3 & L2 plus L3 primary share is 15\%; L1 dominates at 84\% & Not supported as stated \\
\addlinespace
V1b: cross-layer coupling & Failures implicate layers beyond L1 & 82\% of incidents carry secondary codes; 69\% and 83\% of L1 primaries implicate L3 and L4 & Supported, in sharper form \\
\addlinespace
V2: tooling asymmetry & Ecosystem overweight L1 and L4, underweight L2 and L3 & L2 at 0\% Full coverage everywhere; L3 Full only inside orchestration runtimes (100\% there, 0 to 11\% elsewhere) & Supported for L2 outright; for L3 as segregation, not absence \\
\addlinespace
V3a: maturity concentration & Deployments concentrate at L0 to L1 & All 35 in the L0 band; composites 0.00 to 0.13 & Supported, stronger than hypothesized \\
\addlinespace
V3b: modal weakest layer & L3 is the modal weakest layer & L2 weakest or tied weakest in all 35 & Not supported; L2 instead \\
\bottomrule
\end{tabular}
\end{table}

\subsection{A qualifying note on public data inferencing}\label{sec:threats}
Public incident data over-represents spectacular failures and under-represents silent ones, which likely undercounts L4 (silent degradation) and L3 (oversight that failed invisibly). Coding from documentation measures disclosed capability, not deployed capability, in both directions. Single-source coding is mitigated by dual coding and adjudication in Study 1 and by conservative evidence rules in Studies 2 and 3. Two further limits qualify Study 1: the earliest-layer decision rule compresses primary codes toward L1 by construction, relocating the multi-layer signal into the secondary codes, and because large multi-agent estates are only now entering production, the public record under-samples exactly the Layer 2 failure modes that Studies 2 and 3 show the ecosystem is least equipped to observe. The consistency check itself inherits a monoculture caveat by our own Layer 2 logic: two prompt renderings of one model are correlated coders, so $\kappa=0.83$ measures protocol robustness, not coder independence; the cross-provider second coding reported in Section~\ref{sec:validation} and Appendix~\ref{app:protocols} addresses that concern directly, and the lower ($\kappa=0.643$) but still substantial agreement it finds is itself evidence that the two checks measure different things: reduced from the pre-adjudication intra-protocol figure, but reduced by real cross-model disagreement rather than by prompt variation, and the disagreements themselves skewed toward scope-boundary judgment calls (whether a system's behavior qualifies as autonomous multi-step agency) rather than layer misclassification. These limits bound the strength of quantitative claims; the studies are offered as validation of structure, not precise measurement of magnitude.

\section{The CASE Maturity Model}\label{sec:maturity}
\subsection{Levels}
Existing maturity models measure whether governance processes exist \citep{acharya2026, msmaturity2026}. The CASE maturity model measures whether the correct science governs each scale of agency, which makes it a \emph{scientific} maturity model rather than a process one. Five levels are defined by which layers are operational and coupled; see Table~\ref{tab:levels}. The full assessment instrument, anchored scoring bands per layer, a scoring worksheet, and a worked example appear in Appendix~\ref{app:instrument}.

\begin{table}[htbp]
\centering
\small
\caption{CASE maturity levels.}
\label{tab:levels}
\begin{tabular}{lp{2.2cm}p{5.6cm}p{4.2cm}}
\toprule
\textbf{Level} & \textbf{Name} & \textbf{Defining characteristics} & \textbf{Exit criteria} \\
\midrule
L0 & Ad hoc & Prompt-level guardrails only; no observers, no interaction awareness, oversight by informal review; deployment by project teams & First closed-loop agent with audited observability in production \\
\addlinespace
L1 & Controlled & Per-agent closed loops; evaluations as observers; defined intervention surfaces; bounded autonomy per agent & Interaction graph documented and monitored for all multi-agent paths \\
\addlinespace
L2 & Emergence-aware & Interaction-level monitoring; cascade circuit breakers placed by graph analysis; branching factor tracked under load; model-diversity policy on critical paths & Variety inequality (Eq.~\eqref{eq:variety}) computed and closed for every human-agent team \\
\addlinespace
L3 & Requisite & Engineered variety amplification; tiered supervision with handoff contracts; supervisors operate on behavioral models; oversight capacity measured and reported; Article 14 compliant by construction & Autonomy modulated by live error budgets across the fleet (Eq.~\eqref{eq:autonomy}) \\
\addlinespace
L4 & Autonomic & Error-budget-modulated autonomy; zero-touch deployment with policy gates; registry as system of record; self-regulation within certified envelopes; humans govern envelopes, not instances & Sustained operation with coupled-layer certification renewed on change \\
\bottomrule
\end{tabular}
\end{table}

\subsection{A non-compensatory maturity index}\label{sec:index}
Let $m_l \in [0, 1]$ denote the assessed maturity of layer $l$, $\bar{m}$ their arithmetic mean, and $m_{\min} = \min_l m_l$ the binding layer. The composite is the \emph{CASE bottleneck index}, a single score that places majority weight on the binding layer:
\begin{equation}\label{eq:index}
M_{\mathrm{CASE}} = \alpha\, m_{\min} + (1 - \alpha)\, \bar{m}, \qquad \alpha = 0.6
\end{equation}
The maturity level is read directly from the composite: an organization sits at level $k$ when $M_{\mathrm{CASE}}$ falls in the band beginning at $\tau_k$, with $(\tau_1, \tau_2, \tau_3, \tau_4) = (0.25,\, 0.50,\, 0.75,\, 1.00)$ and level L0 below $0.25$. Majority weight on the minimum bounds compensation: with an absent layer ($m_{\min} = 0$) the composite is capped at $0.4\,\bar{m} \leq 0.4$, so no organization with a missing layer can score beyond the low end of the L1 band, and the strict geometric limit, retained below as the robustness case, drives the same organization to exactly zero.

The design is a deliberate consequence of the coupling analysis, not a statistical convenience. Arithmetic averaging, the default in process maturity models, would report an organization with world-class operations and no interaction-layer governance as moderately mature. The strict operationalization of non-compensation is the geometric mean $(\prod_l m_l)^{1/4}$, under which a zero anywhere collapses the composite to zero, exactly as an unmonitored interaction graph nullifies the assurance value of perfect single-agent loops; we retain it as the robustness limit of the index, but as a measurement instrument it is degenerate in exactly the population that matters, since (as Section~\ref{sec:validation} shows) a single universally absent layer maps every organization to the same score. Equation~\eqref{eq:index} resolves the tension: rewriting it as $M_{\mathrm{CASE}} = \bar{m} - \alpha(\bar{m} - m_{\min})$ shows it to be a bounded bottleneck penalty in which the weakest layer carries majority weight, yet organizations remain distinguishable by their capability elsewhere. We also considered a shifted harmonic mean $4 / \sum_l (m_l + \epsilon)^{-1}$, which penalizes low scores smoothly but introduces an arbitrary constant $\epsilon$ precisely where the instrument is most sensitive and assigns a nonzero floor to an organization with no governance at all; we rejected it on both grounds. Reading levels from banded composites preserves the staged pattern practitioners already know from capability maturity models and technology readiness levels, while keeping the instrument a single number. In the scored population the sensitivity result is stronger than a range check: because $m_{L2} = 0$ universally, $M_{\mathrm{CASE}} = (1-\alpha)\,\bar{m}$, so deployment rankings are identical for every $\alpha < 1$ and the conclusions of Section~\ref{sec:validation} hold for any choice of $\alpha$.

For executives, the instrument converts the framework into a capital allocation rule sharper than the geometric version: only investment in the binding layer moves the $m_{\min}$ term, and for the profiles that dominate observed practice it is the only investment that lifts the composite into the next band, so the weakest layer is funded first not by policy preference but by arithmetic. When two layers tie at the minimum, they are funded as a package, since the score moves only when the minimum itself rises. Appendix~\ref{app:instrument} demonstrates this with a worked marginal analysis in which two investments of equal arithmetic value differ materially in both score and resulting level.

\section{Enterprise Illustration: Zero Touch Agent Deployment}\label{sec:illustration}
\begin{figure}[htbp]
\centering
\includegraphics[width=\linewidth]{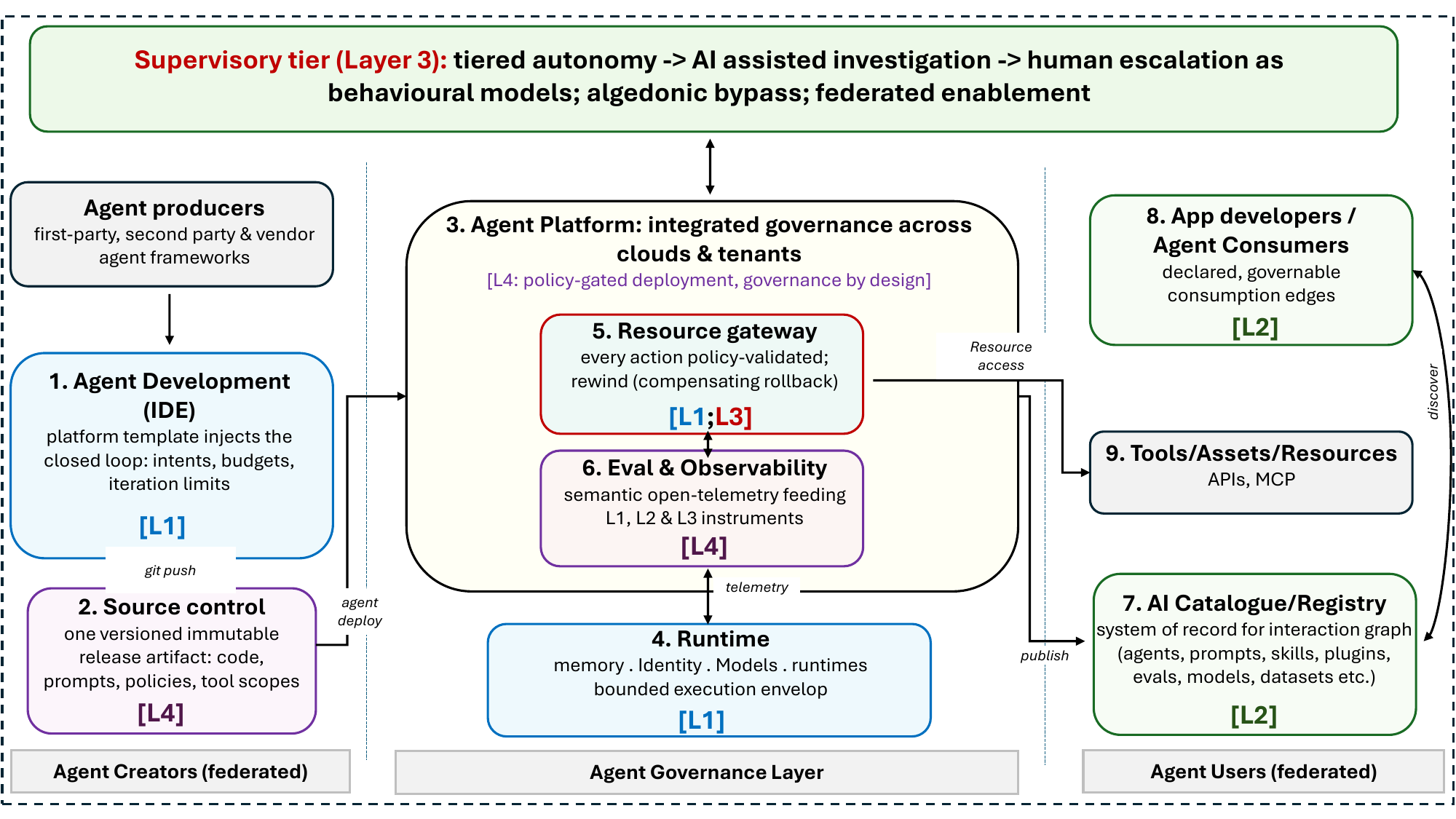}
\caption{Zero Touch Agent Deployment (ZTAD) reference conceptual architecture, redrawn generically from operating platforms. Federated agent producers (left) push versioned release artifacts through a policy-gated governance core (center) onto bounded runtimes, with all resource access routed through centralized resource gateway(s); agents are published to an AI catalogue/registry and consumed by federated application developers (right). Each numbered stage is tinted and tagged by its governing CASE layer: the resource gateway (stage~5) is both a Layer~1 intervention surface and a Layer~3 authority instrument, and the observability plane (stage~6) is the shared substrate that feeds the Layer~1, Layer~2, and Layer~3 instruments. The tools, assets, and resources estate (stage~9) is itself registered and scoped, with resource access granted through the resource gateway. A Layer~3 supervisory tier spans the governance core; the three federated organizational zones appear as the bottom band.}
\label{fig:ztad}
\end{figure}

We illustrate CASE through a reference architecture (Figure~\ref{fig:ztad}) distilled from the operation of production agentic platforms across business functions in more than one hundred markets across large enterprises. Its large-scale deployment model, Zero Touch Agent Deployment (ZTAD), is deliberately more than an automation pipeline: it is the Layer 4 machinery through which the other three layers are enforced by construction, and its lifecycle provides a concrete walk of the framework.

The lifecycle proceeds in nine stages, each mapped to its governing layer. (1)~Agents are developed in standard frameworks by federated producer teams, with platform scaffolding injecting the Layer 1 loop structure: declared intents, budget caps, and iteration limits are part of the template, not an afterthought. (2)~Agent code, prompts, policies, and tool scopes are version-controlled as one release artifact, the precondition for Layer 4 atomic rollback. (3)~Deployment is push- or pull-based through policy-as-code gates that verify observer coverage, declared interaction edges, and least-privilege tool scopes before admission: governance by construction rather than review. (4)~Agents run on selected runtimes with platform-issued memory, cryptographic workload identity, and model bindings, the bounded execution envelope of Layer 1. (5)~All resource access, tools, APIs, and data, flows through a centralized resource gateway (often called a tool gateway) where every action is policy-validated and reversible via compensating rollback (rewind), which is simultaneously a Layer 1 intervention surface and a Layer 3 authority instrument. (6)~All activity is logged into a semantic observability plane over open telemetry standards, the shared substrate of Section~\ref{sec:coupling} that feeds Layer 1 observers, Layer 2 population metrics, and Layer 3 escalation packets. (7)~Agents are published to an AI catalogue and registry, the Layer 2 system of record for the interaction graph, covering agents, prompts, skills, plugins, and loops, both custom and off-the-shelf. (8)~Application developers discover and consume agents through the registry and gateway, which keeps consumption edges declared and therefore governable. (9)~The tools, assets, and resources estate itself (APIs and MCP servers) is registered and scoped, so resource access is granted per agent identity through the resource gateway rather than held ambiently.

Above the lifecycle, supervision is tiered by design: routine variance is absorbed autonomously, structured anomalies route to AI-assisted investigation, and human escalation arrives as a behavioral model (intent, deviation, proposed remediation) rather than raw traces, an explicit variety-amplification chain in the sense of Equation~\eqref{eq:variety}. A federated enablement model trains functional teams as capable supervisors of the agents they own, an implementation of Beer's recursion principle that distributes oversight variety across the organization instead of concentrating it in a central bottleneck.

Three lessons from operating platforms of this class inform the framework directly. First, the coupling of Section~\ref{sec:coupling} is observable in practice: the highest-severity near-incidents involved a Layer 1 observability gap surfacing as a Layer 2 propagation pattern, never a single-layer mechanism in isolation. Second, the zero-touch deployment paradox is not hypothetical: as ZTAD removed the release bottleneck, the binding constraint on scale shifted visibly to Layer 3, the rate at which oversight variety could be engineered, which is precisely what the requisite variety analysis predicts. Third, mechanisms that are not drilled decay: override paths, rollback, and circuit breakers retained their value only when exercised through scheduled fault-injection, confirming the chaos-engineering entry of Table~\ref{tab:mechanisms} as a first-class governance mechanism rather than an operational nicety.

\section{Strategic and Regulatory Implications}\label{sec:implications}
For executives, CASE converts agentic governance from a compliance narrative into an engineering program with a capital allocation rule. The non-compensatory index directs investment to the weakest layer, and the gap diagnosis of Table~\ref{tab:gaps} sharpens that rule into a per-layer agenda (build L2, deepen L3, adopt L4, couple L1); the assessment instrument of Appendix~\ref{app:instrument} generates a board-reportable scorecard; and the framework supplies a common language between risk, technology, and business functions that process maturity models, organized around activities rather than mechanisms, do not. The zero-touch deployment paradox supplies the sharpest budget argument: every dollar of deployment automation creates an oversight liability that must be funded in the same planning cycle, or the organization is buying speed toward the failure of Inequality~\eqref{eq:ztad}.

For regulators and compliance officers, the framework operationalizes oversight requirements. Article 14 of the EU AI Act requires that natural persons can effectively oversee high-risk systems, interpret outputs, and intervene \citep{euaiact}. Inequality~\eqref{eq:variety} is a testable engineering statement of that requirement: an organization that cannot exhibit its variety budget and amplification chain cannot demonstrate effective oversight, whatever its committee structure. We suggest that supervisory authorities and auditors ask for exactly these artifacts, and Table~\ref{tab:mechanisms} names them mechanism by mechanism. The same logic anticipates the direction of the NIST-led agent standards work, which already identifies oversight impossibility without engineered support as a defining property of agentic systems \citep{nist2026}.

For sovereign-scale AI programs, which deploy agent fleets across national institutions and critical services, the stakes of the framework compound: variety deficits at national scale are not reputational risks but institutional ones, and the CASE maturity pathway offers a staging discipline for autonomy expansion that does not depend on trusting any single vendor stack.

For the vendor ecosystem, Study 2 is a quantified roadmap: Layer 2 is an empty quadrant (zero of 22 tools offer Full interaction-graph or cascade monitoring, and only three offer Partial), and Layer 3 capability ships only inside orchestration runtimes rather than in the observability plane where fleets are actually watched. Interaction-graph monitoring and oversight-capacity engineering are the next competitive frontier in agent platforms, and the mechanism inventory of Table~\ref{tab:mechanisms} doubles as the product requirements catalogue for both gaps.

\section{Limitations and Future Work}\label{sec:limitations}
Three limitations bound our claims. First, the formal apparatus is deliberately spare: Equations~\eqref{eq:loop} through \eqref{eq:index} state governance conditions rather than derive closed-form results, and quantities such as the interaction terms $\varphi(i, j)$ and behavioral entropy $H(A)$ are hard to estimate precisely in production; we treat them as measurable in principle and approximable in practice through telemetry, and tighter estimation methods are future work. Second, the empirical studies validate structure from public evidence, with the biases discussed in Section~\ref{sec:threats}; controlled simulation of the coupling dynamics, in the spirit of the validation performed for process maturity models \citep{acharya2026}, is a natural next step. Third, the enterprise illustration distills the operating experience of limited practitioner environments, which demonstrates feasibility, not generality; independent multi-organization field application of the instrument is the appropriate test, and we invite practitioners to apply Appendix~\ref{app:instrument} and report layer profiles.

\section{Conclusion}\label{sec:conclusion}
Agentic AI has revived, at industrial scale and speed, a set of governance problems that older disciplines spent decades solving: keeping an autonomous system on setpoint, containing emergence in interacting populations, matching regulatory variety to system variety, and operating stochastic fleets reliably. The failure of current governance is not a shortage of controls but a category error in their application: the toolkit built for deterministic automation stretched across four scales of stochastic agency. CASE, built by layers individuated by governing science corrects the category error by assigning control theory to the agent, complex adaptive systems theory to the collective, supervisory cybernetics to the human-agent team, and engineering operations to the fleet, coupling them through a shared telemetry substrate, tracing every enterprise control to the classical construct it implements, and binding the whole with a non-compensatory maturity index that makes the weakest layer the binding constraint it truly is. The empirical record gives the weakest layer a name: across every tool and every deployment we could measure, the emergence layer is ungoverned, an Emergence Gap between risk realized and governance deployed, and under coupled failure dynamics an ungoverned layer anywhere is a fragile system everywhere. We have eighty years of science for governing autonomous systems. The work of this decade is not to invent its replacement but to operationalize it, layer by layer, at the scale agentic AI now demands.

\section*{Acknowledgements}
The authors thank Hui (Paul) Xiang (Intel) and Jingwei Zuo (Technology Innovation Institute) for feedback, support and guidance that materially strengthened this paper. We also thank Charles Holive for an insightful discussion in early 2025 on the governance of autonomy, which helped shape the framework presented here. Any remaining errors, and the views expressed, are the authors' own.

\appendix

\section{The CASE Maturity Assessment Instrument}\label{app:instrument}

\subsection{Layer diagnostic rubric}
Table~\ref{tab:rubric} states the diagnostic questions and evidence artifacts per layer. Each layer is scored on the anchored bands of Section~\ref{app:bands}, using documentary evidence only: an undocumented capability scores as absent.

\begin{table}[htbp]
\centering
\small
\caption{Layer diagnostic rubric.}
\label{tab:rubric}
\begin{tabular}{p{2.4cm}p{7.0cm}p{4.4cm}}
\toprule
\textbf{Layer} & \textbf{Diagnostic questions} & \textbf{Evidence artifacts} \\
\midrule
L1 Control & Does every production agent have an explicit setpoint, an observer with audited coverage, and a proven intervention surface? Is guardrail feedback closed-loop rather than allow-or-block? Are saturation (anti-windup) and regime (gain scheduling) handled? & Observer coverage audits; eval suites tied to state space; intervention drill records; guardrail strictness schedules \\
\addlinespace
L2 Adaptive systems & Is the agent interaction graph a managed artifact? Is the branching factor measured under stress? Are shared-memory media governed? Is avalanche-size distribution monitored? Is there a base-model diversity policy? & Interaction graph in registry; cascade stress-test results; breaker placement rationale; shared-memory governance policy \\
\addlinespace
L3 Supervisory & Has peak agent variety been computed per team? What is the amplification chain, and does Eq.~\eqref{eq:variety} hold at peak? Do algedonic bypass alerts exist? Is federated supervision structured on recursion? Is oversight itself measured? & Variety budget calculations; escalation SLAs; algedonic alert definitions; supervisor training and drill records; oversight meta-metrics \\
\addlinespace
L4 Engineering ops & Is there a single platform path to production with policy gates? Are prompts and policies versioned, rollback-tested release artifacts? Is autonomy modulated by a live decision-quality error budget? Are fault-injection drills scheduled? Is toil counted? & Deployment pipeline configuration; error-budget dashboards; rollback and chaos drill records; toil accounting reports \\
\bottomrule
\end{tabular}
\end{table}

\subsection{Anchored scoring bands}\label{app:bands}
Each layer score $m_l$ is assigned on five anchored bands. The bands are defined against the mechanism classes of Table~\ref{tab:mechanisms} for that layer (nine classes at L1, eight at L2, nine at L3, ten at L4), which makes scoring mechanical: assessors count evidence-backed mechanism classes and their deployment breadth, rather than forming holistic judgments.

\begin{table}[htbp]
\centering
\small
\caption{Anchored scoring bands, applied per layer against the mechanism classes of Table~\ref{tab:mechanisms}.}
\label{tab:bands}
\begin{tabular}{lp{11.6cm}}
\toprule
\textbf{Band} & \textbf{Criteria} \\
\midrule
0.00 & No mechanism class of the layer is deployed with documentary evidence. \\
\addlinespace
0.25 & At least one quarter of the layer's mechanism classes are deployed on at least some production agents, with evidence artifacts. Deployment is partial and coverage is not measured. \\
\addlinespace
0.50 & At least half of the layer's mechanism classes are deployed on all critical-path agents, with evidence artifacts and measured coverage. Gaps are known and tracked. \\
\addlinespace
0.75 & All mechanism classes of the layer are deployed fleet-wide, with evidence artifacts, measured coverage, and at least annual drills for every mechanism that has an exercise path (interventions, overrides, rollbacks, fault injection). \\
\addlinespace
1.00 & Band 0.75 conditions hold, and mechanism effectiveness is itself measured (drill timing trends, detection latency, escalation SLA adherence), with coupled-layer certification renewed on every material change. \\
\bottomrule
\end{tabular}
\end{table}

Intermediate scores between bands are permitted when evidence supports them; assessors record the specific mechanism classes counted toward the score so that two assessments are comparable.

\subsection{Scoring worksheet and procedure}
The assessment proceeds in five steps. (1)~\emph{Scope}: enumerate the agent fleet, the multi-agent workflows, and the human-agent teams in scope. (2)~\emph{Evidence collection}: for each layer, gather the artifacts named in Table~\ref{tab:rubric}; interviews may locate artifacts but do not substitute for them. (3)~\emph{Dual scoring}: two assessors independently assign band scores per layer against Table~\ref{tab:bands}, recording counted mechanism classes. (4)~\emph{Adjudication}: divergences of more than one band are resolved by joint evidence review; the adjudicated scores $m_1, \ldots, m_4$ are recorded with rationale. (5)~\emph{Index and report}: compute the composite by Equation~\eqref{eq:index}, read its maturity level from the band it falls in, and identify the binding layer, and state the highest-leverage investment. The recommended cadence is quarterly, aligned with error-budget review, with a full re-assessment on any material platform change.

\subsection{Worked example}
Consider an enterprise with strong single-agent controls and a mature deployment pipeline, moderate supervisory engineering, and early-stage interaction monitoring: adjudicated scores $m_1 = 0.75$, $m_2 = 0.25$, $m_3 = 0.50$, $m_4 = 0.75$. The arithmetic mean reports $\bar{m} = 0.5625$, moderate maturity. The bottleneck composite reports:
\begin{equation*}
M_{\mathrm{CASE}} = 0.6 \times 0.25 + 0.4 \times 0.5625 = 0.375, \qquad \text{composite band: L1}
\end{equation*}
a materially weaker picture, consistent with the coupling analysis: this organization is scaling deployment excellence against a static emergence layer, the zero-touch deployment paradox in scores.

The instrument also discriminates between investments the arithmetic mean cannot tell apart. Raising $m_1$ from $0.75$ to $1.00$ and raising $m_2$ from $0.25$ to $0.50$ each add $0.0625$ to the arithmetic mean. Their effects differ sharply:
\begin{align*}
\text{Invest in L1: } & M_{\mathrm{CASE}} = 0.6 \times 0.25 + 0.4 \times 0.625 = 0.40 \quad (\text{band unchanged: L1}) \\
\text{Invest in L2: } & M_{\mathrm{CASE}} = 0.6 \times 0.50 + 0.4 \times 0.625 = 0.55 \quad (\text{band lifts to L2})
\end{align*}
The binding-layer investment dominates on the score and is the only investment that lifts the level: the capital allocation rule of Section~\ref{sec:maturity} made numerical. When two layers tie at the minimum, the score moves only when both rise, so tied weakest layers are funded as a package; under the geometric robustness limit the same ordering holds.

\section{Coding Protocols for the Empirical Studies}\label{app:protocols}

\subsection{Study 1: incident coding protocol}
\emph{Corpus construction.} Sources: the AI Incident Database, the MIT AI Risk Repository, vendor and platform postmortems, observability-provider field studies, and peer-reviewed failure analyses. Inclusion requires autonomous multi-step behavior (planning plus tool execution), mechanism-level description, and either independent reporting or self-reporting with technical specifics. Exclusions: single-completion model failures without agency, speculative or undocumented accounts, and duplicates across sources (deduplicated by system and date). Two scope rules bound the corpus: physical robotic autonomy (autonomous vehicles, delivery robots) is excluded because the study covers agentic AI in the Section~\ref{sec:lit} sense of systems that plan, reason, and invoke software tools; and controlled benchmark or simulation studies of hypothetical attacks are excluded, while demonstrated exploits against production systems with mechanism-level detail are admissible.

\emph{Decision protocol.} Each incident receives one primary code: L1 if a correctly designed single-agent loop (Table~\ref{tab:mechanisms}, Layer 1 classes) would have prevented it; L2 if the mechanism required inter-agent interaction or shared state; L3 if a human oversight point existed but lacked the capacity, model, or authority to act; L4 if the mechanism was operational (deployment, rollback, cost, monitoring absence). When multiple mechanisms contributed, the primary code is the earliest layer whose correct functioning would have interrupted the trajectory; all other contributing layers are recorded as secondary codes.

\emph{Reliability.} Two machine coding passes forming an intra-protocol consistency check (one model, Claude Opus 4.8, prompted with two differently worded renderings of the same protocol, structured JSON output constrained to the feature schema) extract mechanism features from each candidate's full source text; the layer assignment itself is computed deterministically from the published decision rules, so both coders share a single rule base. Because the API does not expose deterministic sampling, every raw model response is cached and published with the pipeline, making the coding exactly reproducible. Agreement on primary codes is reported as Cohen's kappa before adjudication; all divergences are adjudicated by the first author with written rationale recorded in the published dataset.

This intra-protocol check establishes protocol robustness rather than coder independence, since both passes share one base model. An independent cross-provider check applies the identical, unreworded coder-A pass-1 prompt to a second model from a different provider (GPT-5.6-terra, chosen over the reasoning-optimized flagship tier because this task is single-pass feature extraction against a fixed rule base rather than multi-step inference, confirmed empirically by zero reasoning-token usage on this corpus) across all 363 screened units, yielding Cohen's $\kappa=0.643$ before adjudication (32 disagreements, concentrated in scope-boundary judgment calls, i.e. whether a candidate exhibits autonomous multi-step agentic behavior at all, rather than layer misclassification); mean secondary-code Jaccard overlap on the included incidents is $0.173$. All divergences are adjudicated by the first author with written rationale, following the same procedure as the intra-protocol check; adjudication added 13 incidents the intra-protocol process alone had excluded and reassigned one primary code, and one admission was subsequently reversed on scope review, yielding the $N=62$ corpus of Section~\ref{sec:validation}. Full agreement statistics (confusion matrix, per-layer agreement, disagreement detail) are published alongside the coding dataset.

\emph{Code and data availability.} The collection and coding pipeline (source collectors, deduplication, dual coding, adjudication, and kappa computation) is released as open source at \url{https://github.com/srinivastelukunta/case_framework_arxiv_codes} under the MIT license; the coded datasets, citation ledger, emitted tables, and figure accompany this paper.

\begin{table}[htbp]
\centering
\small
\caption{Study 1 reliability checks and their effect on the published corpus.}
\label{tab:reliability}
\begin{tabular}{p{3.6cm}p{3.3cm}ccp{4.0cm}}
\toprule
\textbf{Check} & \textbf{Coders} & \textbf{$\kappa$} & \textbf{Disagreements} & \textbf{Adjudication outcome} \\
\midrule
Intra-protocol consistency & One model, two prompt renderings (Claude Opus 4.8) & 0.83 & 16 of 363 & All adjudicated; no corpus change \\
\addlinespace
Cross-model independence & Two providers (Claude Opus 4.8 vs GPT-5.6) & 0.643 & 32 of 363 & All adjudicated; 13 candidates admitted (one subsequently reversed on scope review), 1 primary code reassigned ($N$: 50 $\rightarrow$ 62) \\
\bottomrule
\end{tabular}
\end{table}

\subsection{Study 2: tooling capability coding protocol}
\emph{Enumeration.} Tools are enumerated from published systematic mappings \citep{agentops2024}, public repository metadata, and vendor documentation, restricted to tools with public documentation sufficient for capability coding.

\emph{Coding frame.} For each tool and each layer, the coder asks whether the documentation evidences at least one mechanism class of Table~\ref{tab:mechanisms} for that layer: Full (mechanism described with operational detail), Partial (mechanism claimed with limited detail), None. One line of documentation evidence is recorded per non-None cell. Marketing claims without described mechanisms code as None.

\subsection{Study 3: deployment scoring protocol}
\emph{Enumeration.} Publicly describable enterprise deployments are drawn from published case studies, engineering blogs, conference talks, and hiring postings, restricted to deployments with enough disclosure to score at least two layers.

\emph{Scoring.} Each deployment is scored per layer on the bands of Table~\ref{tab:bands} from documentary evidence only, conservative-coded: absence of evidence scores as absence of capability. Hiring signals are admissible as evidence of operated layers (a posted role for oversight engineering evidences L3 investment). Composites are computed by Equation~\eqref{eq:index} and levels read from the composite bands; distributional results are reported with the conservative-coding caveat stated in Section~\ref{sec:threats}.

\end{document}